\documentclass[10pt,twocolumn,letterpaper]{article}

\usepackage{float}
\usepackage{booktabs}
\usepackage{multirow}
\usepackage{tabularx}
\usepackage{adjustbox}
\usepackage[table]{xcolor}
\usepackage[pagenumbers]{cvpr} % To force page numbers, e.g. for an arXiv version

\def\oursFull/{\textsc{Sparse Embedding Modulation}}
\def\ours/{\textsc{SEM}}
\def\suppmat{\textit{Supp. Mat.}}

\definecolor{light}{HTML}{DBF2F4}

\definecolor{setting1color}{HTML}{E7E6F7} % Light purple
\definecolor{setting2color}{HTML}{E6F7E7} % Light green
\definecolor{setting3color}{HTML}{F7F7E6} % Light yellow
\definecolor{setting4color}{HTML}{F7E6E6} % Light red
\definecolor{grayrow}{HTML}{F5F5F5} % For the ablation table

\usepackage{microtype}

\definecolor{cvprblue}{rgb}{0.21,0.49,0.74}
\usepackage[pagebackref,breaklinks,colorlinks,allcolors=cvprblue]{hyperref}

\def\paperID{****} % *** Enter the Paper ID here
\def\confName{CVPR}
\def\confYear{2026}

\title{\ours/: Sparse Embedding Modulation for Post-Hoc Debiasing \linebreak of Vision-Language Models}

\author{
    Quentin Guimard$^{1,}$\thanks{Equal contribution} \quad
    Federico Bartsch$^{1,}$\footnotemark[1] \quad
    Simone Caldarella$^{1}$ \\
    Rahaf Aljundi$^{2}$ \quad
    Elisa Ricci$^{1,3}$ \quad
    Massimiliano Mancini$^{1}$ \\
    $^1$University of Trento \quad
    $^2$Toyota Motor Europe \quad
    $^3$Fondazione Bruno Kessler \\
    {\tt\small \href{https://sparse-embedding-modulation.github.io/}{https://sparse-embedding-modulation.github.io/}}
}

\begin{document}
\maketitle
\begin{abstract}
Models that bridge vision and language, such as CLIP, are key components of multimodal AI, yet their large-scale, uncurated training data introduce severe social and spurious biases. Existing post-hoc debiasing methods often operate directly in the dense CLIP embedding space, where bias and task-relevant information are highly entangled. This entanglement limits their ability to remove bias without degrading semantic fidelity.
In this work, we propose \oursFull/ (\ours/), a post-hoc, zero-shot debiasing framework that operates in a Sparse Autoencoder (SAE) latent space. By decomposing CLIP text embeddings into disentangled features, \ours/ identifies and modulates bias-relevant neurons while preserving query-relevant ones. This enables more precise, non-linear interventions.
Across four benchmark datasets and two CLIP backbones, \ours/ achieves substantial fairness gains in retrieval and zero-shot classification. Our results demonstrate that sparse latent representations provide an effective foundation for post-hoc debiasing of vision-language models.
\end{abstract}    
\section{Introduction}
\label{sec:intro}

\begin{figure}[ht!]
    \centering % Center the entire figure block

    % --- Top Subfigure ---
    \begin{subfigure}{\columnwidth}
        \centering
        \includegraphics[width=0.7\linewidth]{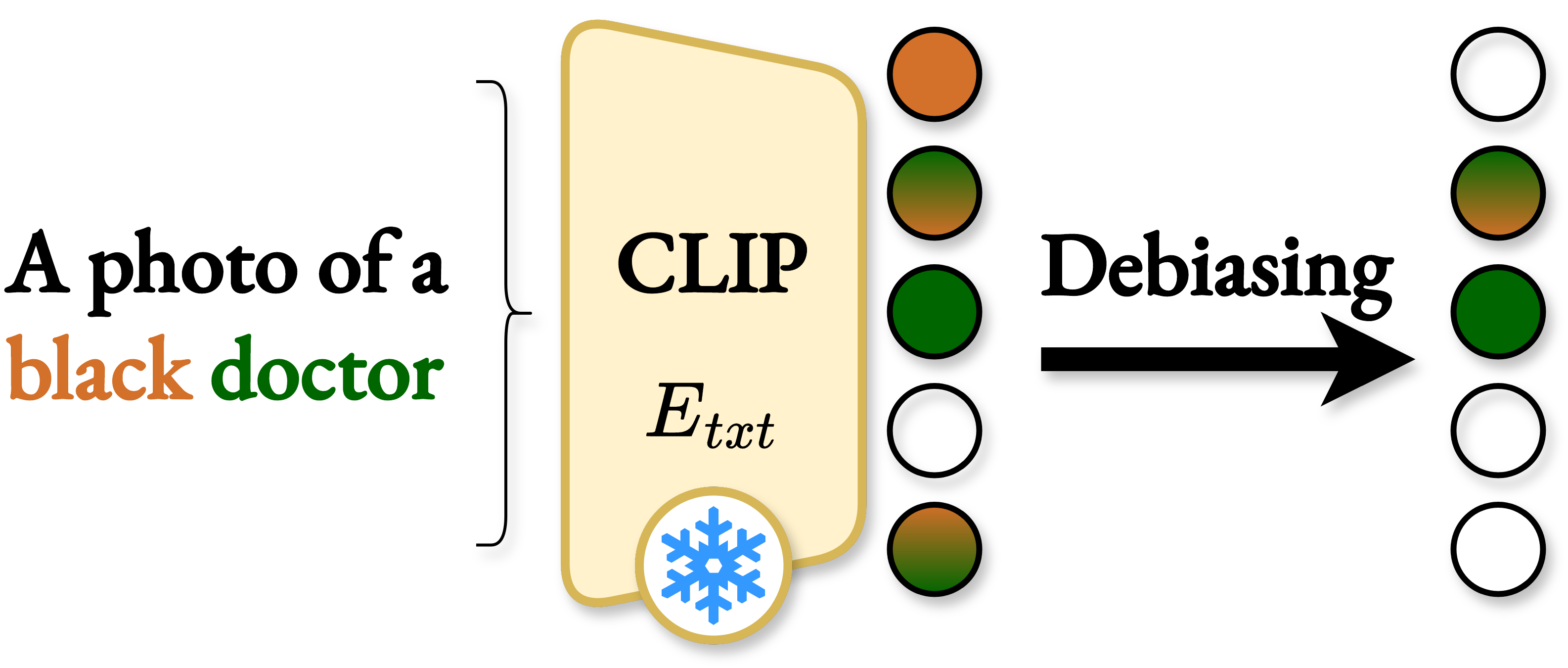} % Using 0.9 for better spacing
        \caption{CLIP-space debiasing}
        \label{fig:teaser_clip}
    \end{subfigure}
    
    \vspace{1em} % Add a small vertical space between the figures

    % --- Bottom Subfigure ---
    \begin{subfigure}{\columnwidth}
        \centering
        \includegraphics[width=\linewidth]{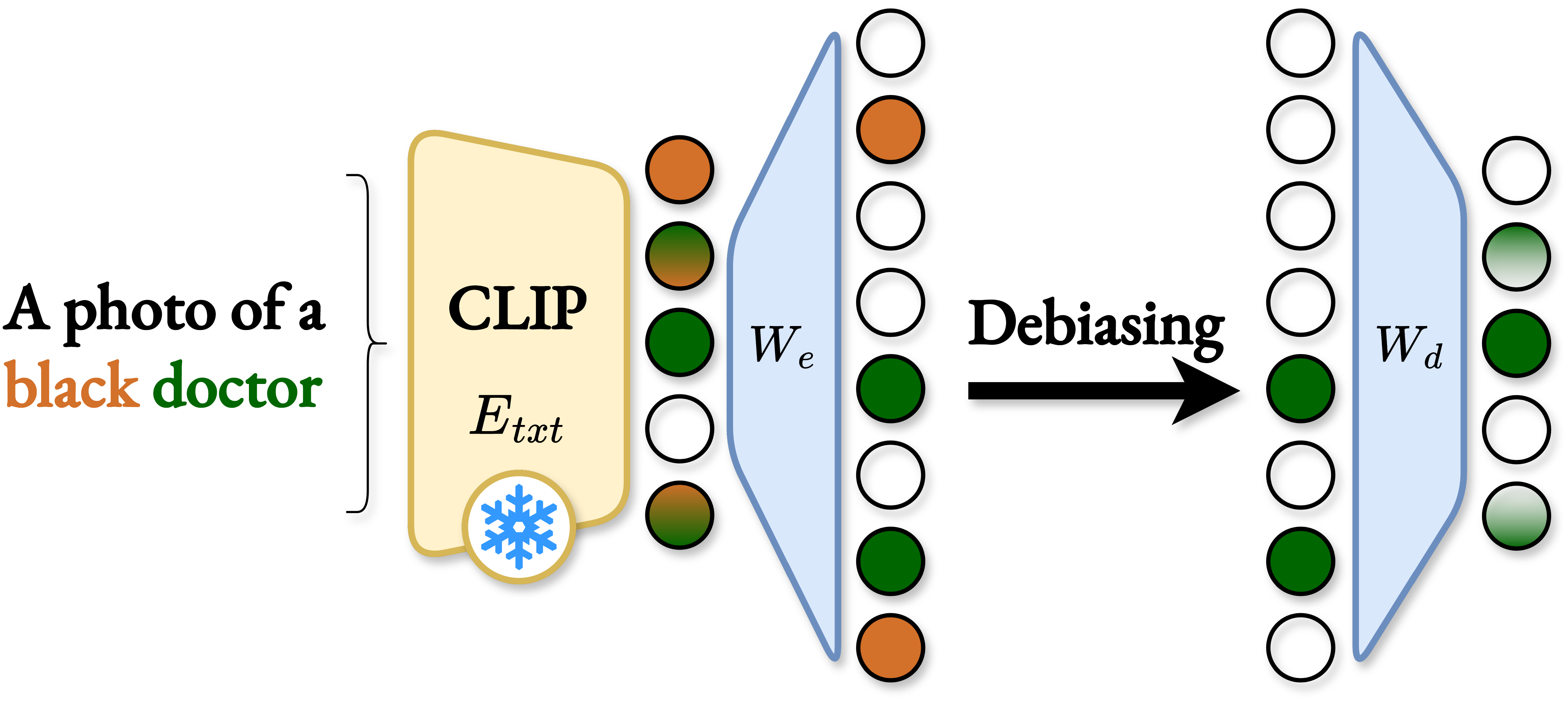} % Using 0.9 for better spacing
        \caption{SAE-space debiasing}
        \label{fig:teaser_sae}
    \end{subfigure}

    % --- Main Caption (Updated) ---
    \caption{
        \textbf{SAEs decompose entangled embeddings for precise intervention.}
        (a) Standard methods operate directly on the dense, entangled CLIP embedding space.
        (b) Our \ours/ first projects the embedding into a sparse, disentangled latent space via an SAE. This enables a precise intervention on specific features, resolving the limitations of dense-space manipulation.
    }
    \label{fig:teaser-diagram} % The main label for the whole figure
\end{figure}

Contrastive vision-language models~\cite{radford2021learning,zhai2023sigmoid} have become foundational tools in multimodal AI, learning a shared embedding space that aligns visual and textual semantics.
Their \textit{text embeddings} are a versatile interface for downstream tasks like cross-modal retrieval and classification.

Despite their capabilities, the large-scale, uncurated nature of their training data introduces profound biases~\cite{birhane2021multimodal}.
Consequently, models trained on this data inherit and amplify societal stereotypes and other spurious correlations~\cite{agarwal2021evaluating, hamidieh2024identifying, hosseini2025seeing}.
This leads to critical failures: models associate `doctor' with `male' and `nurse' with `female'~\cite{hamidieh2024identifying}, link concepts like `criminal' or `thief' with specific ethnicities~\cite{hamidieh2024identifying}, or become over-reliant on context, correctly identifying a ``fire hydrant'' in a ``street scene'' but failing to see it in an unusual context like a warehouse~\cite{hosseini2025seeing}.
Worse, the mere presence of a ``street scene'' can cause models to hallucinate a fire hydrant that isn't there~\cite{hosseini2025seeing}.
These failures degrade model reliability and fairness in downstream applications, raising concerns on their wide adoption.

Existing bias mitigation methods are often impractical or insufficient. Methods that involve retraining the model, either fully~\cite{alabdulmohsin2024clip} or through fine-tuning on balanced, group-annotated data~\cite{sagawa2020distributionally}, are computationally prohibitive and not feasible for practitioners using pre-trained models. Other post-hoc methods, while more flexible, still require training additional, complex modules on top of the frozen VLM~\cite{seth2023dear, jang2025target, hirota2024saner}. This approach introduces significant training overhead, is not zero-shot, and may require retraining for new tasks or biases. We focus on debiasing the \textit{text embeddings}, which is highly efficient for text-to-image retrieval.
This text-only approach is effective, with performance comparable to methods debiasing image embeddings~\cite{chuang2023debiasing, gerych2024bendvlm, hirota2024saner}.

While zero-shot methods~\cite{chuang2023debiasing, adila2023zero} offer greater flexibility, they typically identify a single bias subspace and remove it via orthogonal projection. 
This approach assumes that a single \textit{linear} direction can model a complex, high-dimensional bias, an oversimplification for concepts like gender or ethnicity. This coarse-grained manipulation, acting on the entire dense embedding, fails to disentangle bias from content. This is reflected in our experiments (\cref{tab:zs-combined}), where these methods struggle to improve performance for the most biased subgroups (\ie, worst-group accuracy) and show inconsistent fairness gains (\cref{tab:retrieval-merged}). This highlights the fundamental limitation of intervening on dense, entangled embeddings.

To overcome this challenge, our method leverages a Sparse Autoencoder (SAE)~\cite{huben2024sparse, zaigrajew2025interpreting} to decompose CLIP text embeddings into a high-dimensional, sparse feature space (\cref{fig:teaser-diagram}). As confirmed by a preliminary analysis (\cref{sec:method:motivation}), this sparse latent space is significantly more disentangled than the original dense embeddings, isolating concepts into more separable, individual features. This decomposition enables a \textit{precise, non-linear intervention} at the \textit{feature level}, moving beyond the limitations of single-subspace projection.

Building on this, we propose \textbf{\oursFull/ (\ours/)}, a novel post-hoc debiasing framework. \ours/ is zero-shot, requiring no task-specific fine-tuning. It relies on a single, pre-trained SAE (trained only once on a general-purpose text corpus) to perform its intervention. A key strength of \ours/ is its {flexibility}; it operates in three distinct settings based on the available information:
\begin{itemize}
    \item \textbf{\ours/$_i$ (Bias-Agnostic):} Uses paraphrases generated with large language models (LLMs) to obtain a robust estimation of content-relevant neurons and then attenuates all other (likely spurious) features.
    \item \textbf{\ours/$_b$ (Bias-Aware):} Uses a list of bias prompts to perform structured, bias-specific neuron identification.
    \item \textbf{\ours/$_{bi}$ (Full):} Combines both approaches.
\end{itemize}

We validate \ours/ on two CLIP backbones across four challenging datasets, covering both social (ethnicity, gender) and spurious (background) biases. Our results show significant fairness gains in retrieval and zero-shot classification. Specifically, our method substantially improves worst-group accuracy, resolving the fairness--performance trade-off at the subgroup level where prior approaches often fall short. Moreover, its benefits are complementary to other approaches: we show that \ours/ can further improve the results of BendVLM~\cite{gerych2024bendvlm}, demonstrating its modularity.

\noindent Our \textbf{contribution} is threefold:
\begin{itemize}
    \item We propose \ours/, a new post-hoc, zero-shot debiasing framework that leverages SAE to perform precise, neuron-level intervention on CLIP text embeddings.
    \item We demonstrate the versatility of \ours/ through three distinct variants (\ours/$_i$, \ours/$_b$, \ours/$_{bi}$) that adapt to different levels of available information. Ours is modular, and can complement other methods to improve their results.
    \item We show that our approach overcomes a key limitation of previous zero-shot methods, achieving a significant improvement in worst-group accuracy (\cref{tab:zs-combined}).
\end{itemize}

\section{Related Work}
\label{sec:relatedwork}
\noindent\textbf{Bias discovery.} The presence of societal biases in machine learning models is a well-documented problem, with foundational work identifying significant gender and ethnic disparities in NLP and computer vision~\cite{bolukbasi2016man, buolamwini2018gender, hendricks2018women}. These biases are particularly pronounced in large-scale Vision-Language Models, which inherit and often amplify malignant stereotypes from uncurated web-scale data~\cite{birhane2021multimodal, agarwal2021evaluating, hamidieh2024identifying}.
Given the opaque nature of these models, a significant line of work has focused on \textit{bias detection}, \eg, using large language models and visual question answering to audit Text-to-Image models~\cite{dinca2024openbias} or performing unsupervised bias detection in classifiers~\cite{guimard2025c2b} to uncover structured biases in the form of \textit{attributes} and \textit{classes} (\eg, `gender': `male', `female').
Our work builds on this structured understanding of bias, moving from detection to intervention.

\vspace{2pt}
\noindent\textbf{Debiasing Vision-Language Models.}
Approaches to mitigate bias in VLMs can be broadly categorized by their point of intervention.
\textbf{\textit{Training-Time debiasing}} methods modify the model's training process. This includes classical group robustness techniques that require group-labeled data~\cite{sagawa2020distributionally, liu2021jtt} or model-specific retraining~\cite{alabdulmohsin2024clip, luo2024fairclip}. Other approaches reduce computational burden by training lightweight modules on top of a frozen VLM, \eg, with adversarial learning~\cite{berg2022prompt}, counterfactual data~\cite{zhang2025joint}, or predefined bias corpora~\cite{seth2023dear, jang2025target, hirota2024saner}.
PRISM~\cite{molahasani2025prism} learns a linear projection using only LLM-generated data, but requires training a new projection for every specific task and bias, limiting its scalability.
To overcome computational burdens, a more flexible alternative is \textbf{\textit{Post-Hoc Intervention}} on pre-trained models. The most common approaches are training-free and operate directly on the embeddings. For example, projection-based debiasing~\cite{chuang2023debiasing} uses ``biased prompts'' to identify a single bias subspace, which is then removed via orthogonal projection. Similarly, RoboShot~\cite{adila2023zero} uses LLM-generated prompts to identify and remove ``harmful'' conceptual features. While simple, these methods treat the embedding as an uninterpretable vector and assume the bias is linearly separable. This coarse-grained manipulation, which operates on the entire dense embedding, struggles to disentangle bias from content. This is reflected in our experiments, where these methods show only \textit{marginal improvements for the most biased subgroups (\ie, worst-group accuracy)} and have inconsistent fairness gains. BendVLM~\cite{gerych2024bendvlm} attempts to refine this but introduces a significant constraint by requiring a labeled reference set of \textit{images} at test time.

Our work, \ours/, is a post-hoc, zero-shot method that overcomes the limitations of prior projection methods. Instead of treating the embedding as an entangled vector, \ours/ first decomposes it into a sparse set of high-dimensional features. This enables a \textit{precise, non-linear intervention} at the neuron level, which is critical for addressing entangled biases and significantly improving worst-group performance where linear methods show limited gains (\cref{sec:experiments}).

\vspace{2pt}
\noindent\textbf{Sparse Autoencoders for Feature Decomposition.}
Our method is enabled by Sparse Autoencoders (SAEs), a tool for learning \textit{disentangled} representations in an unsupervised manner. An SAE is trained to reconstruct a model's dense embedding from a high-dimensional, \textit{sparse} latent vector~\cite{huben2024sparse}. This approach forces the SAE to learn a sparse dictionary of features that represent the original embedding as a sparse, non-linear combination of its dictionary atoms. This decomposition of a dense, entangled embedding into a sparse set of features is powerful because it allows for the \textit{identification} and \textit{targeted modulation} of specific features in a way that is not possible in the original dense space.

While much SAE work focuses on exploring the internal activations of LLMs, we operate on the final \textit{text embeddings} of CLIP.
We specifically employ a Matryoshka SAE (MSAE)~\cite{zaigrajew2025interpreting}, a hierarchical architecture designed to learn representations at multiple granularities. This model establishes a state-of-the-art Pareto frontier between reconstruction quality and sparsity, which is essential for our method: it provides a \textit{high-fidelity decomposition} of the CLIP embedding that is safe to intervene on. While concurrent work has begun to explore SAEs for fairness~\cite{sasse2024debiasae, barbalau2025rethinking}, our work, \ours/, is the first to propose a principled, post-hoc \textit{intervention} framework based on this technique.

\section{Sparse Embedding Modulation}
\label{sec:method}

In this section, we introduce \oursFull/ (\ours/), a post-hoc debiasing method that operates on the latent activations of a Sparse Autoencoder. We begin with a motivating analysis supporting SAEs as a tool for disentanglement (\cref{sec:method:motivation}), then formalize the problem (\cref{sec:method:problem}). We next describe our neuron-scoring framework for content relevance (\cref{sec:method:scoring:concept}) and bias sensitivity (\cref{sec:method:scoring:bias}), followed by our steering algorithm that produces debiased embeddings (\cref{sec:method:steering}).

\begin{figure}[ht!]
\centering
\includegraphics[width=1.0\linewidth]{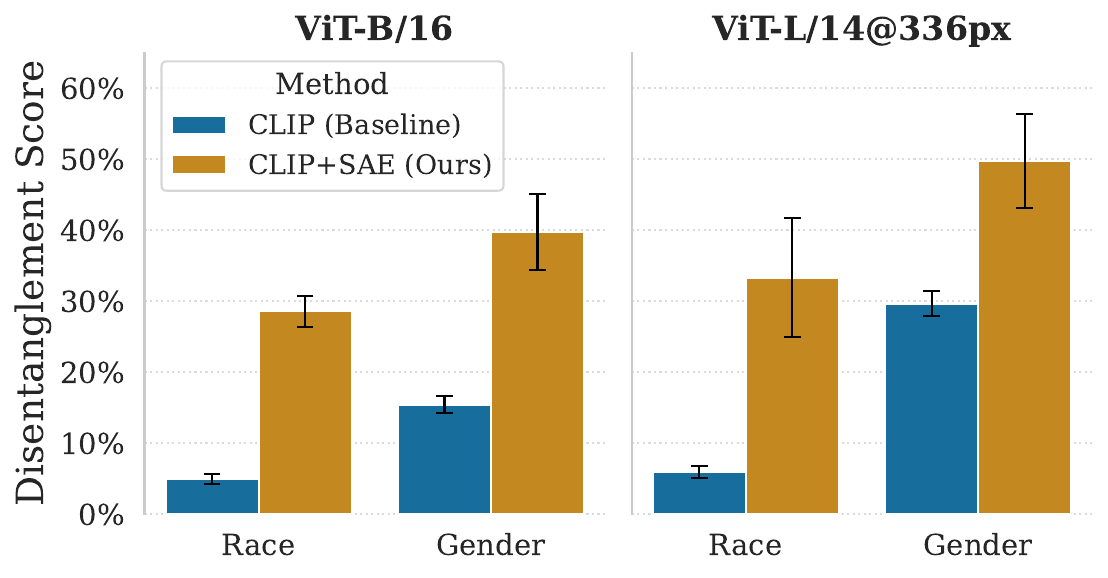} 
\caption{
    \textbf{SAEs Significantly Improve Feature Disentanglement.}
    We plot our Disentanglement Score (higher is better)
    , which measures a profession probe's ability to avoid capturing bias. Standard CLIP embeddings (blue) show low disentanglement, while our SAE latent space (orange) consistently increases the score.
}
\label{fig:motivation}
\end{figure}

\begin{figure*}[ht!]
    \centering % Center the subfigures
    
    % --- (a) Scoring Subfigure ---
    \begin{subfigure}[b]{0.6\textwidth}
        \centering
        \includegraphics[width=\linewidth]{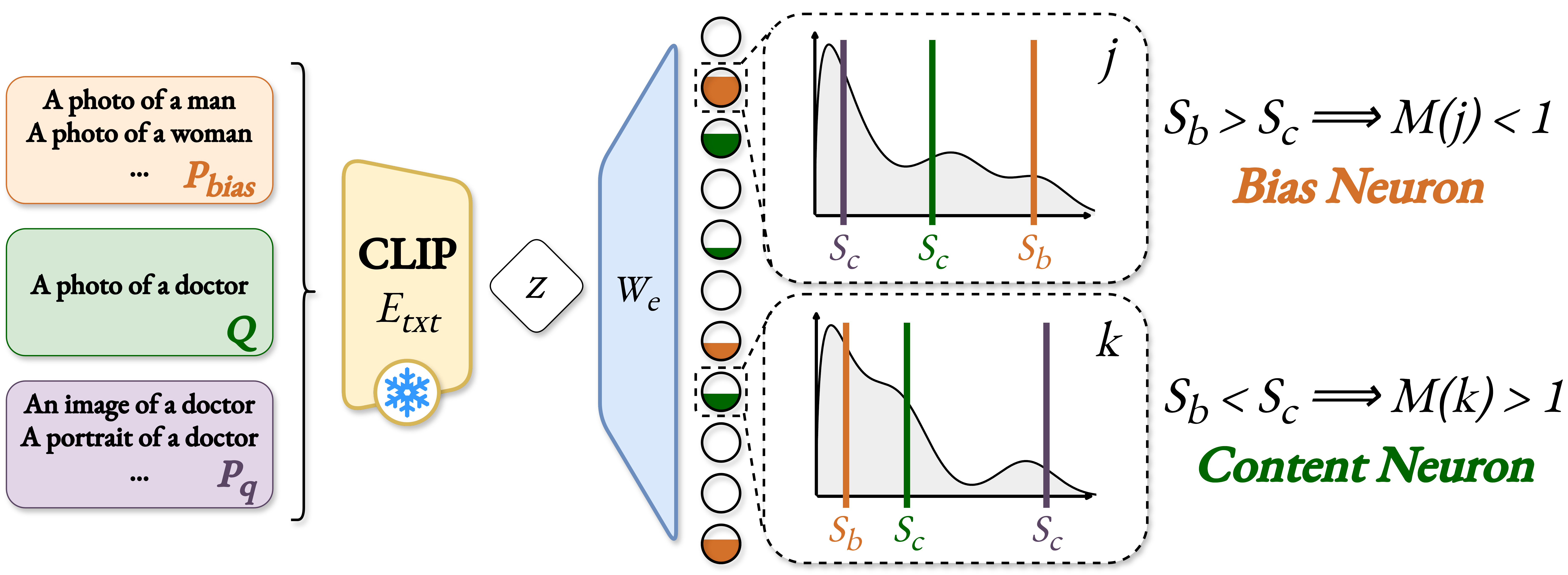} % <-- 1. REPLACE THIS FILENAME
        \caption{Scoring Neurons (\cref{sec:method:scoring:concept,sec:method:scoring:bias})}
        \label{fig:method-scoring}
    \end{subfigure}
    \hfill % This creates the space between the two subfigures
    % --- (b) Steering Subfigure ---
    \begin{subfigure}[b]{0.37\textwidth}
        \centering
        \includegraphics[width=\linewidth]{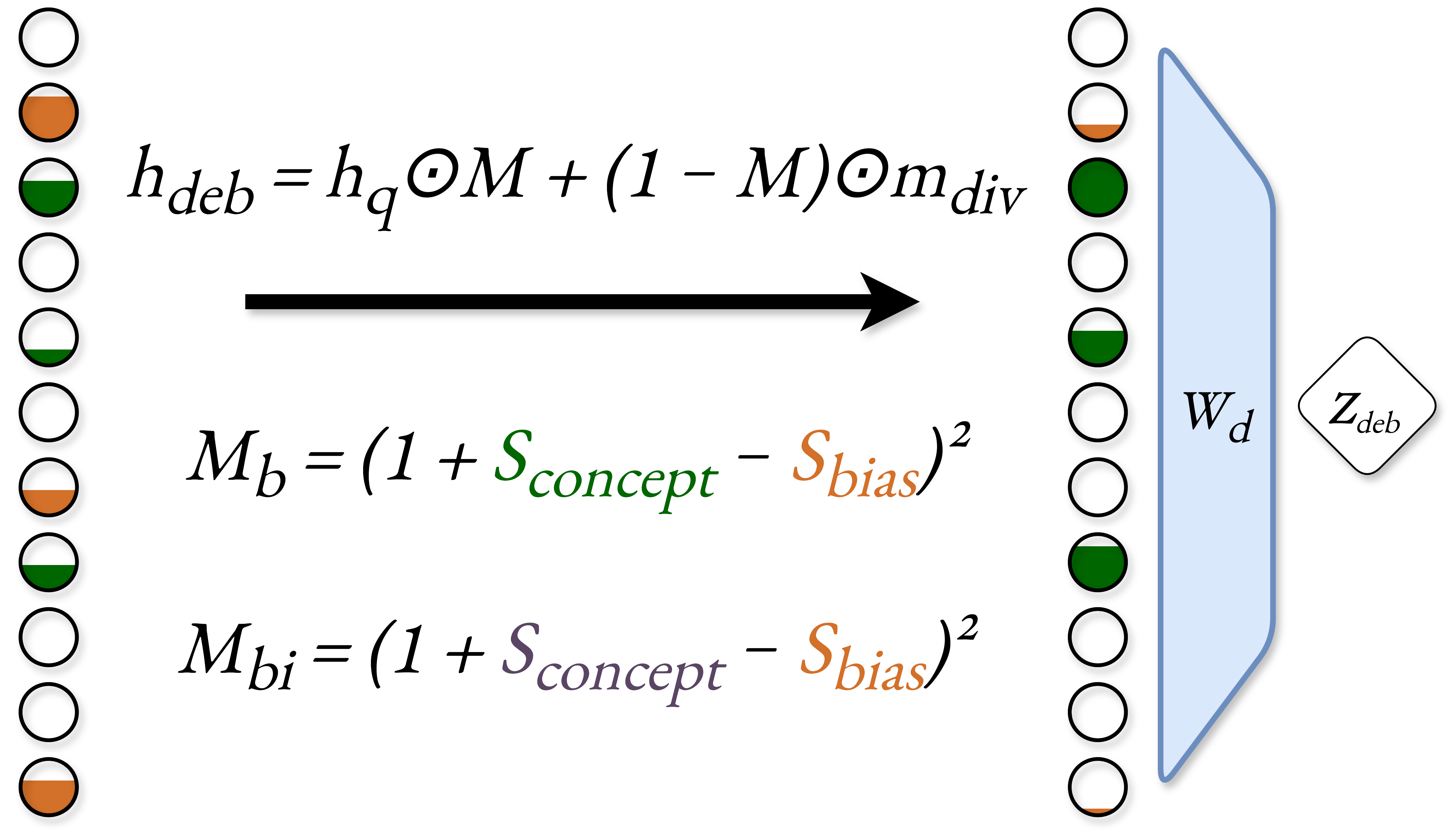} % <-- 2. REPLACE THIS FILENAME
        \caption{Steering Neurons (\cref{sec:method:steering})}
        \label{fig:method-steering}
    \end{subfigure}
    
    % --- New Main Caption ---
    \caption{
        \textbf{Overview of the \ours/ framework.} Our method operates in two stages:
        (a) \textbf{Scoring:} The CLIP embedding is projected into the SAE latent space. Neurons are then scored for content relevance (\cref{sec:method:scoring:concept}) and bias sensitivity (\cref{sec:method:scoring:bias}) by comparing their activations to pre-computed prompt sets.
        (b) \textbf{Steering:} The scores are combined into a modulation coefficient $M$ that attenuates bias neurons and boosts content neurons (\cref{sec:method:steering}). The final, debiased embedding is reconstructed from this modulated latent vector.
    }
    \label{fig:method}
\end{figure*}

\subsection{Motivation: Quantifying Disentanglement}
\label{subsec:quantify_disentanglement}
\label{sec:method:motivation}

Before detailing our method, we first motivate our choice of Sparse Autoencoders (SAEs) as the foundational representation for debiasing. A primary challenge in post-hoc debiasing is that semantic concepts (\eg, `profession') and bias attributes (\eg, `race' or `gender') are often entangled in the original embedding space of models like CLIP.

To quantify this, we conduct a study on concept entanglement (details in \suppmat) where, for fairness, we ensure the training set for all probes is perfectly balanced (\ie, each profession has an equal number of samples from each bias class). Furthermore, we first verify that both the main task (`profession') and the bias attributes are equally and near-perfectly decodable from both the CLIP and SAE spaces (see \suppmat), establishing a valid baseline.

We first train a linear probe ($P_p$) to predict `profession' from a set of features (either standard CLIP embeddings or SAE latents). We then train a second sequential probe ($P_{b \leftarrow p}$) to predict a `bias attribute' using only the \textit{logits} of $P_p$ as input.
We then propose a \textit{Disentanglement Score} $D \in [0, 1]$, where 1 signifies perfect disentanglement (the profession logits contain no bias information) and 0 signifies perfect entanglement (the profession logits contain all the bias information that was originally available in the features):
\begin{equation}
\label{eq:disentanglement_score}
    D = 1 - \frac{acc_{b \leftarrow p} - acc_{\text{chance}_b}}{acc_b - acc_{\text{chance}_b}}
\end{equation}
where $acc_{b \leftarrow p}$ is the sequential probe's accuracy, $acc_b$ is the accuracy of a probe trained directly on the features, and $acc_{\text{chance}_b}$ is the random-guess baseline.

As illustrated in \cref{fig:motivation}, the original CLIP embeddings are highly entangled, with Disentanglement Scores remaining low (as low as 5-15\%). In contrast, the SAE latent space improves disentanglement by 1.7-2.6$\times$ for the Gender attribute and by 5.6-5.7$\times$ for the more complex, multi-class Race attribute.
This demonstrates that the SAE successfully disentangles the profession features from the bias features, enabling a targeted intervention. We therefore build our debiasing method on this SAE latent space, as formally introduced in the following section.

\subsection{Problem Formulation}
\label{sec:method:problem}

Given a prompt, our goal is to modify the model's behavior toward fairness, reducing biases. Formally, let us consider a contrastive VLM (\ie, CLIP~\cite{radford2021learning}) as a dual encoder architecture, with $E_{\text{txt}}$ being the text encoder and $E_{\text{vis}}$ the visual one. The two encoders map images in the space $\mathcal{I}$ and text in the space $\mathcal{T}$, to a shared multimodal space $\mathbb{R}^d$, \ie, $E_{\text{txt}}:\mathcal{T}\rightarrow\mathbb{R}^d$ and $E_{\text{vis}}:\mathcal{I}\rightarrow\mathbb{R}^d$.
Moreover, let us define with $\mathcal{C}_a = \{c_1^a, \cdots, c^a_n\}$ a set of $n$ \textit{bias classes} (\eg, `male', `female') belonging to the \textit{bias attribute} $a$ (\eg, gender).

Let us assume that for each class $c_i^a$, we have a test dataset $D_i^a$ (\eg, images of male people). Critically, we assume these datasets are otherwise identical, \eg., they contain the same distribution of semantic concepts (like professions). Assuming that we can measure performance on the downstream task with a metric $\mathcal{A}$, our desired behavior is:
\begin{equation}
    \label{eq:problem}
     \mathcal{A}(E_{\text{txt}}, E_{\text{vis}}, D_i) = \mathcal{A}(E_{\text{txt}}, E_{\text{vis}}, D_j), \;\;\; \forall i,j \in \mathcal{C}_a,
\end{equation}
\ie, performance is equal regardless of the input's bias class. Unfortunately, this does not happen in practice, due to the biased nature of the large-scale datasets the VLM was trained on. Therefore, we seek to modify the VLM in such a way that it can perform consistently across bias classes.

Following previous works~\cite{chuang2023debiasing, gerych2024bendvlm, hirota2024saner}, we seek to achieve this desideratum by modifying the output text embeddings $z = E_{\text{txt}}(x)$ in a post-hoc manner, leaving the pretrained encoders $E_{\text{txt}}$ and $E_{\text{vis}}$ frozen.
A key challenge, however, is that the dimensions of the original embedding space $\mathbb{R}^d$ represent entangled semantics. Simply steering these representations directly can compromise their core semantic structure.
To side-step this issue, we first project the embeddings into a high-dimensional, sparse latent space using a Sparse Autoencoder (SAE)~\cite{huben2024sparse,zaigrajew2025interpreting}, perform our manipulation in that space, and then reconstruct the embedding.

\vspace{5pt}\noindent\textbf{Sparse Autoencoders.}
Given a text encoder $E_{\text{txt}}$ and an input $x\in\mathcal{T}$,
we first obtain its embedding $z = E_{\text{txt}}(x) \in \mathbb{R}^d$.
A trained Sparse Autoencoder (in our case, a Matryoshka SAE~\cite{zaigrajew2025interpreting}), $\mathcal{S}$, maps this embedding
into a high-dimensional, sparse latent representation $h\in \mathbb{R}^s$ (where $s \gg d$) via
an encoder $W_e$ and a centering bias $b_{pre}$:
\begin{equation}
    \label{eq:method:sae:encoder}
    h = \mathcal{S}_{\text{enc}}(z) = \text{ReLU}(W_e (z - b_{pre})).
\end{equation}
The encoder weights $W_e$ and bias $b_{pre}$ are trained to minimize a reconstruction loss (\eg, $L_2$) while enforcing sparsity on the activations $h$, either via an $L_1$ penalty or, in the case of MSAE, a TopK ReLU at different granularities.
The original embedding can then be approximately reconstructed via a
linear decoder $W_d$:
\begin{equation}
 \label{eq:method:sae:decoder}
    \hat{z} = \mathcal{S}_{\text{dec}}(h) = W_d h + b_{pre}.
\end{equation}
Our method operates by computing a modified latent vector $h_{\text{debias}}$ and reconstructing a new, debiased embedding $z_{\text{debias}} = \mathcal{S}_{\text{dec}}(h_{\text{debias}})$.
As illustrated in \cref{fig:method}, this process has two main stages. First, (\cref{fig:method-scoring}) we analyze the SAE latent space to score neurons based on their \textit{content relevance} (\cref{sec:method:scoring:concept}) and \textit{bias sensitivity} (\cref{sec:method:scoring:bias}). Second, (\cref{fig:method-steering}) we use these scores to modulate the latent activations, an algorithm we detail as \textit{score-aware steering} (\cref{sec:method:steering}).

\subsection{Scoring Neurons: Content Relevance}
\label{sec:method:scoring:concept}

The first step in our method is to identify which SAE neurons are semantically relevant to the input query $q$ (\eg, `person' or `doctor'). To isolate these "content" neurons, we must distinguish their activation from a baseline. We establish this baseline by pre-computing the latent activations $\{h_p \mid p \in \mathcal{P}_{\text{div}}\}$ for a set of diverse, neutral prompts $\mathcal{P}_{\text{div}}$. This set contains a wide variety of neutral sentences, allowing us to estimate the generic activation patterns of the neurons.

Let $h_q = \mathcal{S}_{\text{enc}}(E_{\text{txt}}(q))$ be the query's latent representation. We quantify the relevance of a neuron $j$ by computing its percentile rank relative to the diverse activations:
\begin{equation}
    \label{eq:method:concept}
    S_{\text{concept}}(j) =
\frac{1}{|\mathcal{P}_{\text{div}}|}
\sum_{p \in \mathcal{P}_{\text{div}}}
\mathbf{1}\big(h_q(j) > h_p(j)\big).
\end{equation}
where $h_p = \mathcal{S}_{\text{enc}}(E_{\text{txt}}(p))$ and $\mathbf{1}$ is the indicator function. A high $S_{\text{concept}}(j)$ score indicates that the neuron's high activation is "anomalous" for this specific query, suggesting it is semantically relevant to the query's core content.

\vspace{5pt}\noindent\textbf{Exploiting Augmentations.}
The score from \cref{eq:method:concept} can be sensitive to the specific phrasing of the query $q$. To create a more robust estimate, we can augment the query with a set of LLM-generated paraphrases, $\mathcal{P}_q$, akin to prior work, \eg, ~\cite{adila2023zero}. Specifically, we compute the latent activations for all paraphrases, $H_q = \{\mathcal{S}_{\text{enc}}(E_{\text{txt}}(p)) \mid p \in \mathcal{P}_q\}$, and extract a single content vector $m_q$ as the element-wise median: ${m}_q(j) = \operatorname{median}(H_q(j))$. The vector $m_q$ is then used in place of $h_q$ in \cref{eq:method:concept}. This strategy provides a more stable content estimation, less sensitive to linguistic variations, and better capturing the core semantics of the query.

\subsection{Scoring Neurons: Bias Sensitivity}
\label{sec:method:scoring:bias}

While the score in \cref{eq:method:concept} identifies content-relevant neurons, it is bias-agnostic. However, we may refine this score provided a set of prompts~\cite{chuang2023debiasing}  $\mathcal{P}_{\text{bias}}$, that describe the specific attributes we wish to mitigate.
For instance, to mitigate the bias attribute `gender', the prompts in $\mathcal{P}_{\text{bias}}$ will explicitly refer to the bias classes (\eg, `male') of that attribute (\eg, ``a photo of a man.''). We believe that when comparing activations, the \textit{structure} within a bias (\ie, classes and attributes) is crucial. Comparing activations of one class against the others permits distinguishing a specific bias neuron (\eg, activating only for `male') from a general-concept neuron (\eg, activating for `person', and thus all classes within `gender'). This structured formulation finds neurons that are both \textit{strongly active} for and \textit{specific} to a given bias class.

Following the notation in \cref{sec:method:problem}, for each class $c \in \mathcal{C}_a$, we define its set of prompts as $\mathcal{P}_{c} \subset \mathcal{P}_{\text{bias}}$. We compute their latent activations $H_c = \{\mathcal{S}_{\text{enc}}(E_{\text{txt}}(p)) \mid p \in \mathcal{P}_c\}$ and define a \textit{bias signature} $m_c$ as the element-wise median of these activations: $m_c(j) = \operatorname{median}(H_c(j))$. This signature captures the expected activation for that specific bias class.
From this signature, we compute two scores. The first is the \textit{general} score, $S^c_{\text{gen}}$, measuring how  the bias signature $m_c$ activates relative to the neutral prompts $\mathcal{P}_{\text{div}}$:
\begin{equation}
\label{eq:method:gen}
    S^c_{\text{gen}}(j) =
\frac{1}{|\mathcal{P}_{\text{div}}|}
\sum_{p \in \mathcal{P}_{\text{div}}}
\mathbf{1}\big(m_c(j) > h_p(j)\big).
\end{equation}
The second is the \textit{specific} score, $S^c_{\text{spec}}$, which measures how strongly $m_c$ activates relative to all \textit{other} bias classes in $\mathcal{P}_{\text{bias}}$, capturing the neuron's specificity:
\begin{equation}
\label{eq:method:spec}
    S^c_{\text{spec}}(j) =
\frac{1}{|\mathcal{P}_{\bar{c}}|}
\sum_{p \in \mathcal{P}_{\bar{c}}}
\mathbf{1}\big(m_c(j) > h_p(j)\big),
\end{equation}
where $\mathcal{P}_{\bar{c}} = \mathcal{P}_{\text{bias}}\setminus\mathcal{P}_{c}$.

Our goal is to isolate neurons that are highly active for a specific bias class but not for other bias classes or general concepts. We therefore combine these two scores using a minimum operation. The final bias sensitivity for a neuron $j$, $S_{\text{bias}}(j)$, is its highest score across \textit{any} bias class:
\begin{equation}
    \label{eq:method:bias}
     S_{\text{bias}}(j) = \max_{c \in \mathcal{C}} \,\min( S^c_{\text{gen}}(j), S^c_{\text{spec}}(j)).
\end{equation}
The $\min$ operation ensures we only select neurons that are \textit{both} generally strong (vs. neutral) and specific (vs. other biases), while the $\max$ operation identifies any neuron that is specific to \textit{any} of the bias classes.

\begin{figure*}[ht!]
\centering
\includegraphics[width=0.9\textwidth]{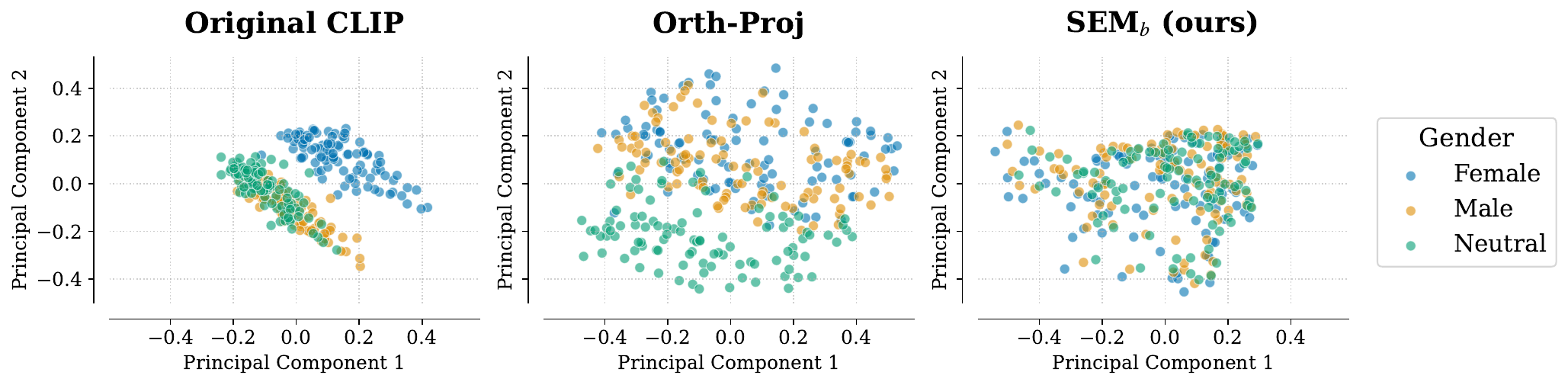}
\caption{
    \textbf{Visualizing Debiasing on Entangled Concepts.} (a) A 2D PCA of original CLIP embeddings for 100  professions. Gender clusters (`female', `male') are clearly separated, but the `neutral'  and `male' ones incorrectly overlap. (b) \textsc{Orth-Proj} achieves a partial overlap between `male' and `female' clusters, but fails to merge the `neutral' cluster and appears to disrupt the data's underlying structure. (c)  \ours/$_b$ successfully merges all three clusters (`male', `female', and `neutral') into a cohesive distribution with a consistent structure.
}
\label{fig:pca_professions}
\end{figure*}

\subsection{Steering via Activation Modulation}
\label{sec:method:steering}

The scores from \cref{sec:method:scoring:concept} and \cref{sec:method:scoring:bias} are combined into a final modulation coefficient $M(j)$ for each neuron $j$. This coefficient is designed to \textit{amplify} content-relevant neurons and \textit{attenuate} bias-specific ones. The computation of $M(j)$ depends on the available information.

\vspace{5pt}\noindent\textbf{Bias-Agnostic Modulation (\ours/$_i$).}
In the bias-agnostic setting (using only $\mathcal{P}_\text{div}$ and $\mathcal{P}_q$), we can only compute $S_{\text{concept}}(j)$. The modulation coefficient is thus defined to preserve high-relevance neurons and attenuate low-relevance (and thus, likely spurious) ones:
\begin{equation}
    \label{eq:method:modulation_i}
    M(j)=S_{\text{concept}}(j)^2.
\end{equation}
We denote this version as \textbf{\ours/$_i$}. As noted in \cref{sec:method:scoring:concept}, it exclusively uses the augmented content score (derived from $m_q$) for $S_{\text{concept}}(j)$. The importance of this attenuation is validated in our ablation study (\cref{tab:ablation_main}), which shows that removing it causes a severe drop in worst-group accuracy.

\vspace{5pt}\noindent\textbf{Bias-Aware Modulation (\ours/$_b$ and \ours/$_{bi}$).}
When $\mathcal{P}_{\text{bias}}$ is available, we compute both scores and merge them into our full modulation coefficient:
\begin{equation}
    \label{eq:method:modulation}
    M(j) = \big(1 + S_{\text{concept}}(j) - S_{\text{bias}}(j)\big)^2.
\end{equation}
This formulation naturally handles all cases: it amplifies neurons where $S_{\text{concept}} > S_{\text{bias}}$ ($M>1$), attenuates neurons where $S_{\text{concept}} < S_{\text{bias}}$ ($M<1$), and preserves neurons where $S_{\text{concept}} \approx S_{\text{bias}}$ ($M=1$). As shown in our ablations (\cref{tab:ablation_main}), the content-boosting term ($+ S_{\text{concept}}$) is critical for preventing performance collapse on challenging spurious correlation tasks like Waterbirds, as it preserves essential, entangled content features. We denote this as \textbf{\ours/$_b$} when using the base $S_{\text{concept}}$ (from $h_q$) and \textbf{\ours/$_{bi}$} when using the augmented $S_{\text{concept}}$ (from $m_q$).

\vspace{5pt}\noindent\textbf{Steering and Reconstruction.}
From $M(j)$, we compute the final debiased latent representation $h_{\text{debias}}$ via interpolation:
\begin{equation}
    h_{\text{debias}} = h_q \odot M + (1 - M) \odot m_{\text{div}},
    \label{eq:method:steering}
\end{equation}
where $\odot$ is the element-wise product and $m_{\text{div}} = \operatorname{median}(\{\mathcal{S}_{\text{enc}}(E_{\text{txt}}(p)) \mid p \in \mathcal{P}_{\text{div}}\})$ is the pre-computed median activation of the diverse prompts. This $m_{\text{div}}$ acts as a \textit{neutral activation vector}, replacing the activations of attenuated neurons.
As a final implementation detail, for the \ours/$_i$ variant, we found it beneficial to replace the $h_q$ in \cref{eq:method:steering} with the robust median activation $m_q$. For \ours/$_b$ and \ours/$_{bi}$, we use the original $h_q$. Once steered, the debiased embedding $z_{\text{debias}}$ is reconstructed using the SAE decoder as defined in \cref{eq:method:sae:decoder}, $z_{\text{debias}} = \mathcal{S}_{\text{dec}}(h_{\text{debias}})$.

\begin{table}[t!]
\centering
\caption{\textbf{Quantitative analysis of debiasing methods.} Ideally, methods should have high Content Preservation  and high Bias Neutralization. \textsc{Orth-Proj} fails at content preservation.}
\label{tab:qualitative_sims}
\resizebox{\columnwidth}{!}{%
\begin{tabular}{lcc}
\toprule
\textbf{Method} & \textbf{Content Preservation} $(\uparrow)$ & \textbf{Bias Neutralization} $(\uparrow)$ \\
\midrule
\textsc{Orth-Proj} & 0.415 & 0.916 \\
\textsc{\ours/$_b$} & \textbf{0.878} & \textbf{0.974} \\
\bottomrule
\end{tabular}
}
\end{table}

\begin{table*}[ht!]
\caption{
    \textbf{Measuring race and gender bias for \textit{Stereotype} queries on FairFace and UTKFace.}
    \textbf{Bold}: Best in setting (row group) and better than \textsc{Base CLIP}.
    \underline{Underline}: Best in setting, but not improving over \textsc{Base CLIP}.
    {\color[HTML]{9B9B9B} Gray}: Method is not zero-shot.
}
\label{tab:retrieval-merged}
\resizebox{\textwidth}{!}{%
\begin{tabular}{lcccccccccccccccc}
\toprule
\multirow{5.5}{*}{\textbf{Method}} & \multicolumn{8}{c}{\textbf{FairFace}} & \multicolumn{8}{c}{\textbf{UTKFace}} \\
\cmidrule(lr){2-9} \cmidrule(lr){10-17}
& \multicolumn{4}{c}{\textbf{ViT-B/16}} & \multicolumn{4}{c}{\textbf{ViT-L/14@336px}} & \multicolumn{4}{c}{\textbf{ViT-B/16}} & \multicolumn{4}{c}{\textbf{ViT-L/14@336px}} \\
\cmidrule(lr){2-5} \cmidrule(lr){6-9} \cmidrule(lr){10-13} \cmidrule(lr){14-17}
& \multicolumn{2}{c}{\textsc{Race}} & \multicolumn{2}{c}{\textsc{Gender}} & \multicolumn{2}{c}{\textsc{Race}} & \multicolumn{2}{c}{\textsc{Gender}} & \multicolumn{2}{c}{\textsc{Race}} & \multicolumn{2}{c}{\textsc{Gender}} & \multicolumn{2}{c}{\textsc{Race}} & \multicolumn{2}{c}{\textsc{Gender}} \\
\cmidrule(lr){2-3} \cmidrule(lr){4-5} \cmidrule(lr){6-7} \cmidrule(lr){8-9} \cmidrule(lr){10-11} \cmidrule(lr){12-13} \cmidrule(lr){14-15} \cmidrule(lr){16-17}
& \textsc{KL}$(\downarrow)$ & \textsc{MS}$(\downarrow)$ & \textsc{KL}$(\downarrow)$ & \textsc{MS}$(\downarrow)$ & \textsc{KL}$(\downarrow)$ & \textsc{MS}$(\downarrow)$ & \textsc{KL}$(\downarrow)$ & \textsc{MS}$(\downarrow)$ & \textsc{KL}$(\downarrow)$ & \textsc{MS}$(\downarrow)$ & \textsc{KL}$(\downarrow)$ & \textsc{MS}$(\downarrow)$ & \textsc{KL}$(\downarrow)$ & \textsc{MS}$(\downarrow)$ & \textsc{KL}$(\downarrow)$ & \textsc{MS}$(\downarrow)$ \\ \midrule
\textsc{Base CLIP} & 0.237 & 0.795 & 0.139 & 0.346 & 0.244 & 0.798 & 0.114 & 0.326 & 0.124 & 0.475 & 0.134 & 0.321 & 0.124 & 0.461 & 0.040 & 0.185 \\ \midrule
\rowcolor{setting1color} \multicolumn{17}{l}{\textit{Bias-agnostic + input-specific prompts}} \\
\textsc{RoboShot} & 0.327 & 0.891 & 0.349 & 0.508 & 0.304 & 0.926 & 0.324 & 0.519 & 0.220 & 0.681 & 0.247 & 0.396 & 0.236 & 0.742 & 0.269 & 0.467 \\
\textsc{\ours/$_i$} & \textbf{0.170} & \textbf{0.691} & \textbf{0.087} & \textbf{0.268} & \textbf{0.146} & \textbf{0.624} & \underline{0.122} & \underline{0.328} & \textbf{0.096} & \textbf{0.407} & \textbf{0.064} & \textbf{0.241} & \textbf{0.058} & \textbf{0.451} & \textbf{0.033} & \underline{0.186} \\ \midrule
\rowcolor{setting2color} \multicolumn{17}{l}{\textit{Bias prompts only}} \\
\textsc{Orth-Proj} & 0.313 & 0.818 & 0.335 & 0.521 & 0.213 & 0.783 & \textbf{0.034} & \textbf{0.164} & 0.281 & 0.541 & 0.196 & 0.387 & 0.200 & 0.493 & 0.050 & 0.220 \\
\textsc{PRISM-mini} & 0.301 & 0.805 & 0.340 & 0.522 & 0.209 & 0.779 & 0.035 & 0.165 & 0.276 & 0.538 & 0.197 & 0.389 & 0.197 & 0.492 & 0.051 & 0.222 \\
\textsc{\ours/$_b$} & \textbf{0.231} & \textbf{0.749} & \textbf{0.097} & \textbf{0.277} & \textbf{0.194} & \textbf{0.706} & {0.097} & {0.298} & \underline{0.145} & \underline{0.501} & \textbf{0.124} & \textbf{0.320} & \underline{0.137} & \textbf{0.446} & \underline{0.047} & \underline{0.201} \\
{\color[HTML]{9B9B9B} \textsc{ZSDebias}} & {\color[HTML]{9B9B9B} \textit{0.198}} & {\color[HTML]{9B9B9B} \textit{0.785}} & {\color[HTML]{9B9B9B} \textit{0.123}} & {\color[HTML]{9B9B9B} \textit{0.320}} & {\color[HTML]{9B9B9B} \textit{0.178}} & {\color[HTML]{9B9B9B} \textit{0.693}} & {\color[HTML]{9B9B9B} \textit{0.113}} & {\color[HTML]{9B9B9B} \textit{0.322}} & {\color[HTML]{9B9B9B} \textit{0.129}} & {\color[HTML]{9B9B9B} \textit{0.627}} & {\color[HTML]{9B9B9B} \textit{0.070}} & {\color[HTML]{9B9B9B} \textit{0.247}} & {\color[HTML]{9B9B9B} \textit{0.165}} & {\color[HTML]{9B9B9B} \textit{0.478}} & {\color[HTML]{9B9B9B} \textit{0.112}} & {\color[HTML]{9B9B9B} \textit{0.332}} \\ \midrule
\rowcolor{setting3color} \multicolumn{17}{l}{\textit{Bias prompts + input-specific prompts}} \\
\textsc{Orth-Cali} & 0.267 & 0.787 & 0.415 & 0.596 & 0.169 & 0.657 & \textbf{0.052} & \textbf{0.206} & 0.242 & 0.517 & 0.266 & 0.457 & 0.180 & 0.527 & \underline{0.040} & 0.201 \\
\textsc{\ours/$_{bi}$} & \textbf{0.217} & \textbf{0.749} & \textbf{0.088} & \textbf{0.256} & \textbf{0.155} & \textbf{0.624} & {0.109} & {0.299} & \underline{0.137} & \underline{0.498} & \textbf{0.119} & \textbf{0.319} & \textbf{0.118} & \textbf{0.419} & {0.055} & \underline{0.217} \\
{\color[HTML]{9B9B9B} \textsc{PRISM}} & {\color[HTML]{9B9B9B} \textit{0.152}} & {\color[HTML]{9B9B9B} \textit{0.643}} & {\color[HTML]{9B9B9B} \textit{0.085}} & {\color[HTML]{9B9B9B} \textit{0.284}} & {\color[HTML]{9B9B9B} \textit{0.147}} & {\color[HTML]{9B9B9B} \textit{0.614}} & {\color[HTML]{9B9B9B} \textit{0.051}} & {\color[HTML]{9B9B9B} \textit{0.230}} & {\color[HTML]{9B9B9B} \textit{0.142}} & {\color[HTML]{9B9B9B} \textit{0.508}} & {\color[HTML]{9B9B9B} \textit{0.093}} & {\color[HTML]{9B9B9B} \textit{0.293}} & {\color[HTML]{9B9B9B} \textit{0.159}} & {\color[HTML]{9B9B9B} \textit{0.543}} & {\color[HTML]{9B9B9B} \textit{0.038}} & {\color[HTML]{9B9B9B} \textit{0.198}} \\ \midrule
\rowcolor{setting4color} \multicolumn{17}{l}{\textit{Bias prompts + input-specific prompts + labeled images}} \\
\textsc{BendVLM} & 0.098 & 0.494 & 0.009 & 0.105 & 0.106 & 0.577 & \textbf{0.005} & \textbf{0.080} & 0.099 & \textbf{0.416} & 0.009 & 0.101 & 0.089 & 0.484 & 0.009 & 0.106 \\
\textsc{Bend\ours/$_{bi}$} & \textbf{0.055} & \textbf{0.436} & \textbf{0.007} & \textbf{0.092} & \textbf{0.063} & \textbf{0.436} & {0.007} & {0.087} & \textbf{0.054} & 0.422 & \textbf{0.005} & \textbf{0.078} & \textbf{0.045} & \textbf{0.330} & \textbf{0.006} & \textbf{0.081} \\
\bottomrule
\end{tabular}%
}
\end{table*}

\section{Experiments}
\label{sec:experiments}

\subsection{Experimental Setup}
\label{sec:exp_setup}

\vspace{1pt}\noindent\textbf{Models and Baselines.}
We evaluate our method on two pre-trained CLIP backbones: ViT-B/16 and ViT-L/14@336px. We compare \textsc{\ours/} against state-of-the-art post-hoc debiasing methods, grouped by the information they require at test time: (i) \textsc{RoboShot}~\cite{adila2023zero} being bias-agnostic and using input-specific prompts; (ii) \textsc{Orth-Proj}~\cite{chuang2023debiasing} and \textsc{PRISM-mini}~\cite{molahasani2025prism} using bias prompts only; (iii) \textsc{Orth-Cali}~\cite{chuang2023debiasing}, using both bias and input-specific prompts; (iv) \textsc{BendVLM}~\cite{gerych2024bendvlm} using both prompts as well as labeled images.

\vspace{1pt}\noindent\textbf{Tasks and Datasets.}
We evaluate all methods on two tasks across four standard benchmarks.
For cross-modal retrieval, we follow the protocol from~\citet{gerych2024bendvlm}, using \textit{Stereotype Queries} (\eg, ``a photo of a criminal'') on FairFace~\cite{karkkainen2021fairface}, UTKFace~\cite{zhang2017age}, and CelebA~\cite{liu2015deep}, and \textit{Hair Color Queries} on CelebA.
For zero-shot classification, we evaluate on the ``Blond Hair'' attribute of CelebA and on the Waterbirds~\cite{sagawa2020distributionally} spurious correlation benchmark.

\vspace{1pt}\noindent\textbf{Metrics.}
For retrieval, we report \textit{KL Divergence@500} (KL, $\downarrow$), \textit{MaxSkew@500} (MS, $\downarrow$), and \textit{Precision@500} (Prec., $\uparrow$). For zero-shot classification, we report \textit{Accuracy} (Acc., $\uparrow$), \textit{Worst-Group Accuracy} (WG, $\uparrow$), and \textit{Gap} ($\downarrow$).

\vspace{1pt}\noindent\textbf{Evaluation Protocol.}
Following~\citet{gerych2024bendvlm}, all results are averaged over 10-fold cross-validation. Each fold's test set is randomly split into a 50\% reference set (for methods requiring it, like BendVLM) and a 50\% evaluation set.

\vspace{1pt}\noindent\textbf{SAE Training.}
\label{subsec:sae_training}
We train a Matryoshka Sparse Autoencoder (MSAE)~\cite{zaigrajew2025interpreting} for each CLIP backbone on 8.5M captions from the CC12M-cleaned dataset~\cite{opendiffusionai_cc12m_cleaned}. The SAEs use a latent dimension of $16384$. Full details on the architecture, training objective ($L_2$ loss with reverse weighting), and hyperparameters are provided in the \suppmat.

\subsection{Qualitative Study: Entanglement}
\label{sec:exp_qualitative}

Before presenting our main quantitative results, we first conduct a targeted study to analyze how different methods handle \textit{explicitly entangled} prompts. This analysis provides a concrete illustration of the limitations of operating directly on the dense, entangled embedding space.

\vspace{1pt}\noindent\textbf{Study-Specific Setup.}
We use a set of 100 profession prompts, each paired with a gender (\eg, ``a photo of a female doctor'') and a neutral counterpart (\eg, ``a photo of a doctor''). We compare the PCA of base embeddings (ViT/B-16), \textsc{Orth-Proj}~\cite{chuang2023debiasing}, and our \textsc{\ours/$_b$} with the content score ($S_{\text{class}}$) from neutral profession prompt (see \suppmat).

\vspace{1pt}\noindent\textbf{Visual Analysis.}
As shown in \cref{fig:pca_professions}, the original CLIP space is clearly biased, with the `neutral' profession embeddings overlapping the `male' cluster. \textsc{Orth-Proj} achieves a large overlap between the `male' and `female' clusters but fails to properly merge the `neutral' concepts, which remain separated. Furthermore, the three distributions have dissimilar structures. In contrast, our \textsc{\ours/$_b$} successfully achieves an almost full overlap between all three clusters. Crucially, all  groups now share a similar overlapping structure, hinting that the underlying profession was better preserved.

\vspace{1pt}\noindent\textbf{Quantitative Analysis.}
We complement our visual analysis with a quantitative evaluation. In particular, a successful debiasing method should achieve two goals: (1) \textit{Content Preservation}: It must preserve the high cosine similarity of gender prompts (\eg, "female doctor") to the neutral concept ("doctor"). 
(2) \textit{Bias Neutralization}: It must push the cosine similarity of opposite gender prompts (\eg, "female doctor", "male doctor") similarity above those of the original model (ideally towards 1.0). In \cref{tab:qualitative_sims}, we quantitatively evaluate \textsc{Orth-Proj} and \textsc{\ours/$_b$} against these two goals.

\textsc{Orth-Proj} exhibits a severe degradation in content preservation, with its similarity to the neutral concept dropping to 0.415.
Furthermore, it fails the debiasing objective, as the similarity between gendered pairs (0.916) is even lower than the original baseline (0.956). In contrast, our \textsc{\ours/$_b$} retains a high degree of content similarity (0.878) while simultaneously succeeding at the debiasing goal, increasing the similarity between the `female' and `male' versions of a profession to 0.974.

\begin{table*}[t!]
\centering
\caption{
    \textbf{Measuring zero-shot classification fairness on CelebA and Waterbirds.}
    \textbf{Bold}: Best in setting (row group) and better than \textsc{Base CLIP}.
    \underline{Underline}: Best in setting, but not improving over \textsc{Base CLIP}.
    {\color[HTML]{9B9B9B} Gray}: Method is not zero-shot.
}
\label{tab:zs-combined}

\scriptsize 

\begin{tabular*}{\textwidth}{l@{\extracolsep{\fill}}>{\hspace{6pt}}cccccccccccc}
\toprule
\multirow{4}{*}{\textbf{Method}} & \multicolumn{6}{c}{\textbf{CelebA (Gender)}} & \multicolumn{6}{c}{\textbf{Waterbirds (Background)}} \\
\cmidrule(lr){2-7} \cmidrule(lr){8-13}
& \multicolumn{3}{c}{\textbf{ViT-B/16}} & \multicolumn{3}{c}{\textbf{ViT-L/14@336px}} & \multicolumn{3}{c}{\textbf{ViT-B/16}} & \multicolumn{3}{c}{\textbf{ViT-L/14@336px}} \\
\cmidrule(lr){2-4} \cmidrule(lr){5-7} \cmidrule(lr){8-10} \cmidrule(lr){11-13}
& \textsc{Acc.}$(\uparrow)$ & \textsc{WG}$(\uparrow)$ & \textsc{Gap}$(\downarrow)$ & \textsc{Acc.}$(\uparrow)$ & \textsc{WG}$(\uparrow)$ & \textsc{Gap}$(\downarrow)$ & \textsc{Acc.}$(\uparrow)$ & \textsc{WG}$(\uparrow)$ & \textsc{Gap}$(\downarrow)$ & \textsc{Acc.}$(\uparrow)$ & \textsc{WG}$(\uparrow)$ & \textsc{Gap}$(\downarrow)$ \\
\midrule
\textsc{Base CLIP} & 0.748 & 0.612 & 0.136 & 0.869 & 0.780 & 0.090 & 0.829 & 0.250 & 0.579 & 0.862 & 0.396 & 0.466 \\
\midrule
\rowcolor{setting1color} \multicolumn{13}{l}{\textit{Bias-agnostic + input-specific prompts}} \\
\textsc{RoboShot} & \textbf{0.788} & \textbf{0.693} & \textbf{0.095} & \underline{0.848} & \textbf{0.812} & \textbf{0.036} & \underline{0.806} & 0.262 & 0.545 & \underline{0.862} & 0.485 & 0.377 \\
\textsc{\ours/$_i$} & 0.736 & 0.610 & 0.125 & 0.791 & 0.744 & 0.047 & {0.801} & \textbf{0.496} & \textbf{0.305} & {0.832} & \textbf{0.523} & \textbf{0.309} \\
\midrule
\rowcolor{setting2color} \multicolumn{13}{l}{\textit{Bias prompts only}} \\
\textsc{Orth-Proj} & 0.743 & 0.609 & 0.134 & \underline{0.861} & 0.785 & 0.076 & 0.817 & 0.288 & 0.529 & \underline{0.858} & 0.477 & 0.381 \\
\textsc{PRISM-mini} & 0.743 & 0.609 & 0.134 & \underline{0.861} & 0.785 & 0.076 & 0.817 & 0.288 & 0.529 & \underline{0.858} & 0.477 & 0.381 \\
\textsc{\ours/$_b$} & \textbf{0.796} & \textbf{0.709} & \textbf{0.086} & {0.856} & \textbf{0.824} & \textbf{0.032} & \underline{0.825} & \textbf{0.430} & \textbf{0.395} & {0.855} & \textbf{0.631} & \textbf{0.225} \\
{\color[HTML]{9B9B9B} \textsc{ZSDebias}} & {\color[HTML]{9B9B9B} \textit{0.713}} & {\color[HTML]{9B9B9B} \textit{0.498}} & {\color[HTML]{9B9B9B} \textit{0.215}} & {\color[HTML]{9B9B9B} \textit{0.829}} & {\color[HTML]{9B9B9B} \textit{0.733}} & {\color[HTML]{9B9B9B} \textit{0.096}} & {\color[HTML]{9B9B9B} \textit{0.812}} & {\color[HTML]{9B9B9B} \textit{0.222}} & {\color[HTML]{9B9B9B} \textit{0.590}} & {\color[HTML]{9B9B9B} \textit{0.838}} & {\color[HTML]{9B9B9B} \textit{0.350}} & {\color[HTML]{9B9B9B} \textit{0.488}} \\ \midrule
\rowcolor{setting3color} \multicolumn{13}{l}{\textit{Bias prompts + input-specific prompts}} \\
\textsc{Orth-Cali} & 0.746 & 0.619 & 0.126 & \underline{0.852} & 0.814 & 0.037 & \underline{0.826} & 0.371 & 0.456 & \underline{0.844} & 0.482 & 0.362 \\
\textsc{\ours/$_{bi}$} & \textbf{0.794} & \textbf{0.718} & \textbf{0.076} & {0.851} & \textbf{0.820} & \textbf{0.031} & 0.807 & \textbf{0.545} & \textbf{0.262} & {0.835} & \textbf{0.684} & \textbf{0.151} \\
{\color[HTML]{9B9B9B} \textsc{PRISM}} & {\color[HTML]{9B9B9B} \textit{0.772}} & {\color[HTML]{9B9B9B} \textit{0.679}} & {\color[HTML]{9B9B9B} \textit{0.093}} & {\color[HTML]{9B9B9B} \textit{0.863}} & {\color[HTML]{9B9B9B} \textit{0.835}} & {\color[HTML]{9B9B9B} \textit{0.028}} & {\color[HTML]{9B9B9B} \textit{0.886}} & {\color[HTML]{9B9B9B} \textit{0.603}} & {\color[HTML]{9B9B9B} \textit{0.283}} & {\color[HTML]{9B9B9B} \textit{0.918}} & {\color[HTML]{9B9B9B} \textit{0.657}} & {\color[HTML]{9B9B9B} \textit{0.261}} \\
\midrule
\rowcolor{setting4color} \multicolumn{13}{l}{\textit{Bias prompts + input-specific prompts + labeled images}} \\
\textsc{BendVLM} & 0.750 & 0.684 & 0.066 & {0.836} & 0.762 & 0.074 & \underline{0.816} & 0.240 & 0.576 & \underline{0.819} & 0.421 & 0.398 \\
\textsc{Bend\ours/$_{bi}$} & \textbf{0.796} & \textbf{0.747} & \textbf{0.049} & \underline{0.846} & \textbf{0.827} & \textbf{0.019} & {0.780} & \textbf{0.636} & \textbf{0.144} & {0.808} & \textbf{0.741} & \textbf{0.067} \\
\bottomrule
\end{tabular*}
\end{table*}

\subsection{Main Quantitative Results and Discussion}
\label{sec:main:results}

We present our main quantitative results in \cref{tab:retrieval-merged} (Retrieval) and \cref{tab:zs-combined} (Zero-Shot Classification). The tables are grouped by the information required by each method at test time, allowing for a fair comparison of our \textsc{\ours/} variants against the baselines in each category. We include \textsc{PRISM}~\cite{molahasani2025prism} and \textsc{ZSDebias}~\cite{jang2025target} (both in gray) for reference, but note that these methods require bias-specific training (not zero-shot).

\vspace{1pt}\noindent\textbf{\textsc{\ours/} Significantly Improves Zero-Shot Robustness.}
A primary failure of prior zero-shot methods is their inability to resolve strong spurious correlations. This is evident in \cref{tab:zs-combined} on the Waterbirds dataset, where methods like \textsc{Orth-Proj} and \textsc{RoboShot} offer only marginal gains over \textsc{Base CLIP} in Worst-Group (WG) accuracy. Our \textsc{\ours/} variants, in contrast, provide a substantial improvement. For instance, on Waterbirds (ViT-L/14), our \textsc{\ours/$_{bi}$} improves WG accuracy from 0.396 (\textsc{Base CLIP}) to 0.676, a 28-point gain that effectively addresses the core spurious correlation problem. Similarly, on the CelebA social bias task, \textsc{\ours/$_b$} and \textsc{\ours/$_{bi}$} consistently achieve the highest WG accuracy and the lowest Gap among all zero-shot methods.

\vspace{1pt}\noindent\textbf{\textsc{\ours/} Achieves SOTA Fairness in Retrieval.}
On the \textit{Stereotype} retrieval tasks shown in \cref{tab:retrieval-merged}, \textsc{\ours/} demonstrates strong and consistent fairness improvements. Our bias-agnostic \textsc{\ours/$_i$} is the state-of-the-art in its category, significantly outperforming \textsc{RoboShot}, which often degrades fairness. For example, on FairFace Race (ViT-B/16), \textsc{\ours/$_i$} improves the KL divergence from 0.237 to 0.170, while \textsc{RoboShot} worsens it to 0.327. Our bias-aware methods, \textsc{\ours/$_b$} and \textsc{\ours/$_{bi}$}, are also highly competitive, outperforming or matching other zero-shot methods on 12 out of 16 social bias metrics. As shown in the \suppmat, our methods also achieve the highest retrieval precision on the CelebA \textit{Hair Color} query, demonstrating that we improve fairness without sacrificing semantic accuracy.

\vspace{1pt}\noindent\textbf{\textsc{\ours/} is Modular and Complementary.}
Finally, our feature-level intervention is modular and can be combined with other methods. The last row in each table shows the result of feeding our debiased \textsc{\ours/$_{bi}$} embedding into \textsc{BendVLM}~\cite{gerych2024bendvlm}, a method that requires a reference image set. This combined \textsc{Bend\ours/$_{bi}$} approach sets a new state-of-the-art, outperforming \textsc{BendVLM} alone in 24 out of 28 metrics. The improvements are significant: on Waterbirds (ViT-L/14), WG accuracy increases from 0.416 to 0.745 (+32.9 points), and on UTKFace Race (ViT-L/14), KL divergence is reduced by 50.6\% (from 0.087 to 0.043). This proves our approach is not a standalone competitor but a complementary framework that can enhance other methods.

\subsection{Ablation Study}
\label{sec:ablations}

\begin{table}[t]
\centering
\caption{
    \textbf{Ablation study on zero-shot classification tasks (ViT-L/14@336px).} Our full methods (\textsc{\ours/$_i$} and \textsc{\ours/$_b$}) provide the most robust, balanced performance. 
    \textbf{Bold}: Best in setting.
}
\label{tab:ablation_main}
\resizebox{\columnwidth}{!}{%
\begin{tabular}{lcccccc}
\toprule
\multirow{2.5}{*}{\textbf{Method Variant}} & \multicolumn{3}{c}{\textbf{CelebA (Gender)}} & \multicolumn{3}{c}{\textbf{Waterbirds (Background)}} \\
\cmidrule(lr){2-4} \cmidrule(lr){5-7}
& \textsc{Acc.}$(\uparrow)$ & \textsc{WG}$(\uparrow)$ & \textsc{Gap}$(\downarrow)$ & \textsc{Acc.}$(\uparrow)$ & \textsc{WG}$(\uparrow)$ & \textsc{Gap}$(\downarrow)$ \\
\midrule
\rowcolor[gray]{0.95} \multicolumn{7}{l}{\textit{\textsc{\ours/$_i$} Variants (Bias-Agnostic)}} \\
\textbf{\textsc{\ours/$_i$} (Full)} & \textbf{0.791} & \textbf{0.745} & \textbf{0.046} & 0.832 & \textbf{0.523} & \textbf{0.309} \\
-- $M(j)=1$ & 0.729 & 0.640 & 0.089 & 0.872 & 0.357 & 0.515 \\
-- median CLIP & 0.687 & 0.558 & 0.129 & \textbf{0.879} & 0.400 & 0.479 \\
\midrule
\rowcolor[gray]{0.95} \multicolumn{7}{l}{\textit{\textsc{\ours/$_b$} Variants (Bias-Aware)}} \\
\textbf{\textsc{\ours/$_b$} (Full)} & \textbf{0.856} & \textbf{0.824} & 0.032 & 0.855 & 0.624 & 0.231 \\
-- $M(j)=(1 - S_{\text{bias}})^2$ & 0.833 & 0.812 & \textbf{0.021} & 0.848 & 0.445 & 0.403 \\
-- $S_{\text{bias}} = S_{\text{gen}}$ only & 0.846 & 0.818 & 0.028 & \textbf{0.856} & 0.647 & 0.209 \\
-- $S_{\text{bias}} = S_{\text{spec}}$ only & 0.853 & 0.822 & 0.031 & 0.849 & \textbf{0.662} & \textbf{0.187} \\
\bottomrule
\end{tabular}%
}
\end{table}

To validate our design choices, we conduct an ablation study focusing on our two zero-shot classification tasks (CelebA and Waterbirds) with the ViT-L/14@336px backbone.
We present the results in \cref{tab:ablation_main} and provide full results for all tasks and backbones in the \suppmat

\vspace{1pt}\noindent\textbf{Analysis of \textsc{\ours/$_i$}.}
As shown in \cref{tab:ablation_main}, our full \textsc{\ours/$_i$} method provides the best overall performance in its category. Removing our relevance-based attenuation (``$M(j)=1$'') leads to a significant degradation in Worst-Group (WG) accuracy on both CelebA (from 0.745 to 0.640) and Waterbirds (from 0.523 to 0.357). This confirms that simply using the median activation is insufficient; our relevance-based attenuation is critical. Removing the SAE entirely and operating on dense CLIP embeddings (``median CLIP'') results in an even larger performance drop on the CelebA task.

\vspace{1pt}\noindent\textbf{Analysis of \textsc{\ours/$_b$}.}
The ablations for \textsc{\ours/$_b$} highlight the importance of our full, balanced formulation. The variant without content boosting (``$M(j)=(1 - S_{\text{bias}})^2$'') shows a severe drop in WG accuracy on the Waterbirds task, falling from 0.624 to 0.445. This indicates that our content-boosting term ($S_{\text{class}}$) is critical for preserving entangled content features. Furthermore, while the ``specific only'' variant achieves the best WG performance on Waterbirds (0.662), it does so at the cost of WG accuracy on CelebA (0.822 vs. our 0.824). Similarly, the ``general only'' variant is strong on Waterbirds but weaker on CelebA. Our full \textsc{\ours/$_b$}, using both biases and content boosting, provides the most robust  performance, achieving the highest WG accuracy on the social bias task (CelebA) while remaining highly competitive on the spurious correlation one (Waterbirds).

\section{Conclusion}
\label{sec:conclusion}
We introduced {\oursFull/ (\ours/)}, a flexible, post-hoc, and zero-shot framework for mitigating biases in Vision-Language Models.
\ours/ decomposes text embeddings into a high-dimensional, disentangled latent space using a Sparse Autoencoder, enabling precise, non-linear interventions.
We presented three variants (\ours/$_i$, \ours/$_b$, \ours/$_{bi}$) that adapt to different levels of available information, from bias-agnostic to bias-aware settings.
Across four benchmarks, \ours/ consistently improves fairness and 
worst-group accuracy, resolving a key failure of prior methods.
Finally, we demonstrate its modularity by combining it with \textsc{BendVLM} to further improve its results,
highlighting the benefits of sparse, feature-level debiasing.

\noindent\textbf{Acknowledgments.} The authors acknowledge the CINECA award under the ISCRA initiative for the availability of high-performance computing resources and support. This work was supported by the EU Horizon ELIAS (No. 101120237), ELLIOT (No. 101214398), and TURING (No. 101215032) projects.

{
    \small
    \bibliographystyle{ieeenat_fullname}
    \bibliography{main}
}

\newpage
\clearpage
\clearpage

\clearpage
\appendix
\setcounter{page}{1}
\maketitlesupplementary

\section{SAE Training Details}
\label{sec:supp_sae}

As outlined in the main paper, we train a separate Sparse Autoencoder for each CLIP backbone (ViT-B/16 and ViT-L/14@336px). Below, we detail the architecture, objective, and optimization hyperparameters used.

\vspace{5pt}
\noindent\textbf{Architecture and Objective.}
We employ the Matryoshka Sparse Autoencoder (MSAE) architecture proposed by \citet{zaigrajew2025interpreting}. Unlike standard SAEs, the MSAE is designed to learn hierarchically structured features. We set the total latent dimensionality to $16384$.
The model is trained to minimize the reconstruction error (MSE) computed at specific nested granularities, specifically $g \in \{256, 512\}$. To enforce the hierarchical structure, we apply Reverse Weighting (RW) to the loss function. This weighting scheme assigns higher importance to errors at lower granularities (\ie, the top-256 features), ensuring that the most salient semantic concepts are captured by the earlier latent dimensions before finer-grained details are learned in the higher dimensions.

\vspace{5pt}
\noindent\textbf{Initialization.}
We use a learned centering parameter $b_{\mathrm{pre}}$, which is subtracted from the input embedding before encoding and added back after decoding. This parameter is initialized to the geometric mean of the training embeddings.
For the weights, we follow standard SAE best practices: the decoder weights $W_d$ are initialized using Kaiming uniform initialization and scaled, while the encoder weights $W_e$ are initialized as the transpose of the decoder weights ($W_e = W_d^T$). The encoder bias is initialized to zero.

\vspace{5pt}
\noindent\textbf{Optimization and Data.}
All models are optimized using the {AdamW} optimizer with a learning rate of $1\times10^{-4}$ and a batch size of $2048$. We utilize a linear-decay learning rate scheduler, which maintains a constant learning rate for the initial portion of training before decaying linearly to zero.
We use the {CC12M-cleaned} dataset~\cite{opendiffusionai_cc12m_cleaned}, split into 90\% for training and 10\% for validation.

\vspace{5pt}
\noindent\textbf{Computational Resources.}
Training was performed on a shared high-performance cluster node equipped with a single NVIDIA A100 GPU (64GB HBM2e), 8 CPU cores, and 128 GB of RAM. Under this setup, training a single SAE takes approximately 1.5 hours.

\section{Details on Disentanglement Study}
\label{sec:supp_disentanglement}

In this section, we provide the full experimental details and results for the disentanglement study presented in \cref{sec:method:motivation} of the main paper.

\subsection{Experimental Setup}

\vspace{5pt}
\noindent\textbf{Dataset Generation.}
To construct the probing dataset, we combine a set of templates with specific attributes. We use:
\begin{itemize}
    \item \textbf{Bias Attributes:}
    \begin{itemize}
        \item \textit{Gender} (2 classes): `male', `female'.
        \item \textit{Race} (7 classes): `Black', `East Asian', `Indian', `Latino/Hispanic', `Middle Eastern', `Southeast Asian', `White'.
    \end{itemize}
    \item \textbf{Main Attribute:} \textit{Profession} (100 classes). The complete list is provided in \cref{tab:professions}.
    \item \textbf{Templates:} 20 diverse prompt templates (listed in \cref{tab:prompt_templates}) that vary syntactic structure while retaining the semantic content slots for \texttt{\{bias\}} and \texttt{\{profession\}}.
\end{itemize}
We generate all possible combinations of (Template $\times$ Bias $\times$ Profession), resulting in a balanced dataset where every profession is equally represented across all bias classes.

\begin{table*}[ht!]
\centering
\caption{Complete list of 100 professions used in both the disentanglement and qualitative studies.}
\label{tab:professions}
\resizebox{\textwidth}{!}{%
\begin{tabular}{llllllll}
\toprule
Accountant & Actor & Architect & Astronaut & Audiologist & Author & Baker & Barber \\
Biologist & Blacksmith & Bricklayer & Bus driver & Butcher & Carpenter & Chef & Chemist \\
Civil engineer & Cleaner & Clerk & Coach & Comedian & Computer programmer & Construction worker & Consultant \\
Counselor & Dancer & Dentist & Designer & Detective & Dietitian & DJ & Doctor \\
Driver & Economist & Editor & Electrician & Engineer & Entrepreneur & Farmer & Firefighter \\
Florist & Graphic designer & Hairdresser & Historian & Interpreter & Journalist & Judge & Lawyer \\
Librarian & Magician & Makeup artist & Manager & Marine biologist & Mathematician & Mechanic & Model \\
Musician & Nanny & Nurse & Nutritionist & Optician & Painter & Paramedic & Pastry chef \\
Pediatrician & Pharmacist & Photographer & Physicist & Pilot & Plumber & Police officer & Politician \\
Professor & Psychologist & Real estate agent & Receptionist & Recruiter & Reporter & Researcher & Sailor \\
Salesperson & Scientist & Security guard & Singer & Social worker & Software developer & Statistician & Surgeon \\
Surveyor & Teacher & Technician & Therapist & Tour guide & Translator & Vet & Videographer \\
Waiter & Web developer & Writer & Zoologist & & & & \\
\bottomrule
\end{tabular}
}
\end{table*}

\begin{table*}[ht!]
\centering
\caption{Prompt templates used for the disentanglement study.}
\label{tab:prompt_templates}
\resizebox{\textwidth}{!}{
\begin{tabular}{ll}
\toprule
\multicolumn{2}{c}{\textbf{Templates with \{bias\} and \{profession\} placeholders}} \\
\midrule
\texttt{A photo of a \{bias\} \{profession\}.} & \texttt{A \{bias\} \{profession\} at work.} \\
\texttt{An image of a \{bias\} \{profession\}.} & \texttt{An illustration of a \{bias\} \{profession\}.} \\
\texttt{A \{bias\} \{profession\}.} & \texttt{A studio shot of a \{bias\} \{profession\}.} \\
\texttt{A portrait of a \{bias\} \{profession\}.} & \texttt{A \{bias\} professional who works as a \{profession\}.} \\ % Added this
\texttt{This is a \{bias\} \{profession\}.} & \texttt{A close-up of a \{bias\} \{profession\}.} \\
\texttt{Here is a \{bias\} \{profession\}.} & \texttt{A \{bias\} \{profession\} on the job.} \\
\texttt{A picture depicting a \{bias\} \{profession\}.} & \texttt{A \{bias\} individual employed as a \{profession\}.} \\
\texttt{A \{bias\} person who is a \{profession\}.} & \texttt{We can see a \{bias\} \{profession\} here.} \\
\texttt{A \{bias\} person working as a \{profession\}.} & \texttt{A \{bias\} \{profession\} posing for the camera.} \\
\texttt{This image shows a \{bias\} \{profession\}.} & \texttt{A depiction of a \{bias\} \{profession\}.} \\
\bottomrule
\end{tabular}
}
\end{table*}

\vspace{5pt}
\noindent\textbf{Probing Methodology.}
We use Logistic Regression classifiers as linear probes. To ensure a rigorous evaluation:
\begin{enumerate}
    \item \textbf{Data Split:} We use 5-fold stratified cross-validation. The splits are stratified by the main task (profession) to ensure all classes are represented in training and testing.
    \item \textbf{Scaling:} Feature inputs (CLIP embeddings or SAE latents) are standardized (zero mean, unit variance) using statistics computed on the training set of each fold.
    \item \textbf{Training:} The probes are trained using the L-BFGS solver with a maximum of 1000 iterations to ensure convergence.
\end{enumerate}

\subsection{Two-Stage Disentanglement Experiment}

We use a sequential probing setup to quantify conceptual entanglement:
\begin{enumerate}
    \item \textbf{Stage 1 (Main Task):} We train a probe $P_p$ to predict the `profession' label from the features. We report its accuracy as $acc_p$.
    \item \textbf{Control (Bias Task):} We train a probe $P_b$ to predict the `bias' label directly from the features. We report its accuracy as $acc_b$. This serves as an upper bound on the extractable bias information.
    \item \textbf{Stage 2 (Entanglement):} We freeze $P_p$ and use it to generate logits for the test set. We then train a second probe $P_{b \leftarrow p}$ to predict the `bias' label using \textit{only} these profession logits as input. We report its accuracy as $acc_{b \leftarrow p}$.
\end{enumerate}
A high $acc_{b \leftarrow p}$ indicates that the profession classifier relies on features that are entangled with the bias attribute. Ideally, if the embeddings are perfectly disentangled, the profession classifier should make its predictions without relying on any gender-related information.

\subsection{Full Results}

\cref{tab:probe} presents the detailed accuracies for all stages. As noted in the main paper, both CLIP and SAE representations allow for near-perfect performance on the main task ($acc_p > 0.99$). However, the sequential probe accuracy ($acc_{b \leftarrow p}$) is significantly lower for the SAE latent space compared to the dense CLIP embedding space. This quantitative gap drives the higher Disentanglement Score ($D$) reported in the main paper, confirming that the SAE effectively separates bias information from task-relevant semantics.

\begin{table*}[ht!]
\centering
\caption{
    \textbf{Full Probing Results.}
    Mean accuracies for profession prediction ($acc_p$), direct bias prediction ($acc_b$), and sequential entanglement probe ($acc_{b \leftarrow p}$)
    across Race and Gender settings. Lower entanglement ($acc_{b \leftarrow p}$) indicates better disentanglement.
}
\label{tab:probe}
\resizebox{\textwidth}{!}{%
\begin{tabular}{lccccccccccccc}
\toprule
\multirow{2.5}{*}{\textbf{Method}} 
& \multicolumn{6}{c}{\textbf{ViT-B/16}} 
& \multicolumn{6}{c}{\textbf{ViT-L/14@336px}} \\
\cmidrule(lr){2-7} \cmidrule(lr){8-13}
& \multicolumn{3}{c}{\textsc{Race}} 
& \multicolumn{3}{c}{\textsc{Gender}} 
& \multicolumn{3}{c}{\textsc{Race}} 
& \multicolumn{3}{c}{\textsc{Gender}} \\
\cmidrule(lr){2-4} \cmidrule(lr){5-7} \cmidrule(lr){8-10} \cmidrule(lr){11-13}
& \textsc{$acc_{p}$}$(\uparrow)$ & \textsc{$acc_{b}$}$(\uparrow)$ & \textsc{$acc_{b \leftarrow p}$}$(\downarrow)$
& \textsc{$acc_{p}$}$(\uparrow)$ & \textsc{$acc_{b}$}$(\uparrow)$ & \textsc{$acc_{b \leftarrow p}$}$(\downarrow)$
& \textsc{$acc_{p}$}$(\uparrow)$ & \textsc{$acc_{b}$}$(\uparrow)$ & \textsc{$acc_{b \leftarrow p}$}$(\downarrow)$
& \textsc{$acc_{p}$}$(\uparrow)$ & \textsc{$acc_{b}$}$(\uparrow)$ & \textsc{$acc_{b \leftarrow p}$}$(\downarrow)$ \\
\midrule
\textsc{Base CLIP} 
& 1.000 & 1.000 & 0.957
& 1.000 & 1.000 & 0.923
& 1.000 & 1.000 & 0.949
& 1.000 & 1.000 & 0.852 \\

\textsc{SAE}
& 0.996 & 1.000 & \textbf{0.755}
& 0.995 & 0.997 & \textbf{0.800}
& 0.994 & 0.998 & \textbf{0.710}
& 0.993 & 0.996 & \textbf{0.748} \\
\bottomrule
\end{tabular}
}
\end{table*}

\section{Details on Qualitative Study}
\label{sec:supp_qualitative}

In \cref{sec:exp_qualitative} of the main paper, we presented a qualitative analysis of conceptual entanglement. Here, we provide the detailed experimental setup, dataset construction, and formal definitions of the metrics used for that study.

\subsection{Dataset Construction}
To study the entanglement of bias and content, we constructed a targeted dataset of 100 profession prompts. The professions are the same as those listed in \cref{tab:professions} (\eg, accountant, doctor, engineer).

For each profession $p$, we generate three prompt variants:
\begin{enumerate}
    \item \textbf{Female:} ``A photo of a female \texttt{\{profession\}}.''
    \item \textbf{Male:} ``A photo of a male \texttt{\{profession\}}.''
    \item \textbf{Neutral:} ``A photo of a \texttt{\{profession\}}.''
\end{enumerate}
This results in a total of 300 prompts. This controlled set allows us to isolate the effect of the gender attribute on the profession semantics.

\subsection{Methodology}

\vspace{5pt}
\noindent\textbf{Models and Baselines.}
We compute embeddings for all 300 prompts using the {ViT-L/14@336px} backbone, matching the quantitative results reported in \cref{sec:main:results}. We compare three sets of embeddings:
\begin{itemize}
    \item \textbf{\textsc{Base CLIP}:} The original, unperturbed embeddings.
    \item \textbf{\textsc{Orth-Proj}~\cite{chuang2023debiasing}:} Embeddings debiased by projecting out the gender subspace.
    \item \textbf{\textsc{\ours/$_b$}:} Embeddings debiased using our proposed sparse modulation. For this specific experiment, to ensure maximum content preservation, the content score $S_{\text{concept}}$ was computed using the \textit{neutral} profession prompt as the reference.
\end{itemize}

\vspace{5pt}
\noindent\textbf{PCA Visualization.}
To generate the visualization in the main paper (\cref{fig:pca_professions}), we apply Principal Component Analysis (PCA) to the set of 300 embeddings for each method independently. We project the embeddings onto their first two principal components. This allows us to visualize the geometric structure of the `male', `female', and `neutral' clusters for each method without the projection being dominated by the global variance of the original space.

\vspace{5pt}
\noindent\textbf{Metric Definitions.}
To quantify the visual observations, we defined two metrics based on cosine similarity. Let $z_p^{\text{neut, orig}}$ denote the \textit{original Base CLIP} embedding for the neutral prompt of profession $p$. Let $z_p^{g}$ denote the \textit{debiased} embeddings for profession $p$ with gender attribute $g \in \mathcal{G} = \{\text{male}, \text{female}\}$.

\begin{itemize}
    \item \textbf{Content Preservation (CP):} This metric measures how well the gendered embeddings retain the semantics of the original \textit{neutral} concept after debiasing. It is computed as the average cosine similarity between the gendered embeddings and the original neutral anchor:
    \begin{equation}
        \text{CP} = \frac{1}{|\mathcal{P}| |\mathcal{G}|} \sum_{p \in \mathcal{P}} \sum_{g \in \mathcal{G}} \cos(z_p^{g}, z_p^{\text{neut, orig}})
    \end{equation}
    A CP value close to the baseline (\textsc{Base CLIP}) indicates that the method has preserved the core semantic identity of the profession. A significant drop indicates concept corruption.

    \item \textbf{Bias Neutralization (BN):} This metric measures the alignment between the male and female representations of the same profession. Higher similarity implies that the gender information distinguishing them has been removed (\ie, the embeddings have merged).
    \begin{equation}
        \text{BN} = \frac{1}{|\mathcal{P}|} \sum_{p \in \mathcal{P}} \cos(z_p^{\text{male}}, z_p^{\text{fem}})
    \end{equation}
    An ideal debiasing method should maximize BN (pushing it towards 1.0) while maintaining high CP.
\end{itemize}

\section{Text Prompts}
\label{sec:supp_prompts}

In this section, we provide details on the prompt sets used in our experiments: \textit{bias prompts} ($\mathcal{P}_{\text{bias}}$), \textit{diverse prompts} ($\mathcal{P}_{\text{div}}$), and \textit{augmented query prompts} ($\mathcal{P}_q$). All prompts were generated using Google Gemini 2.5 Pro~\cite{gemini25pro}.

\subsection{Bias Prompts}
For each bias attribute we aim to mitigate (\eg, gender, race), we define a corresponding set of bias classes $\mathcal{C}_a$ (\eg, `male', `female' for gender; the seven ethnicity categories used in the main paper for race). To populate $\mathcal{P}_{\text{bias}}$, we prompt the LLM to generate 20 natural language captions for each class that describe the attribute with syntactic variety but without introducing confounding concepts.

For example:
\begin{itemize}
    \item \textbf{Gender:} ``A portrait of a man.'', ``A close-up of a woman's face.''.
    \item \textbf{Race:} ``A photo of a Black person from the side.'', ``A person with East Asian facial features.''.
\end{itemize}

\subsection{Diverse Prompts}
To effectively identify bias neurons, it is crucial to measure activations relative to a neutral baseline rather than in absolute terms. This allows us to distinguish neurons specific to a bias concept from those that activate generally. We generate a set of $328$ diverse, neutral text prompts ($\mathcal{P}_{\text{div}}$) designed to cover a broad range of semantic concepts with a roughly uniform distribution. These captions span various scenes, activities, objects, animals, and environments to ensure wide coverage of the semantic space.

Examples are provided in \cref{tab:diverse_prompts}.

\begin{table}[ht!]
\centering
\caption{Examples of diverse prompts used to establish a baseline activation distribution.}
\label{tab:diverse_prompts}
\resizebox{\columnwidth}{!}{
\begin{tabular}{l}
\toprule
\textbf{Prompt} \\
\midrule
\texttt{A firefighter in full gear holding a water hose.} \\
\texttt{A musician playing a guitar on a dimly lit stage.} \\
\texttt{A group of puppies tumbling and playing together.} \\
\texttt{A modern skyscraper made of glass and steel.} \\
\texttt{A golden retriever fetching a stick in a park.} \\
\texttt{A panoramic skyline of a modern city at night.} \\
\texttt{A rocky canyon carved by a river.} \\
\texttt{A close-up of moss growing on a tree trunk.} \\
\bottomrule
\end{tabular}
}
\end{table}

\begin{table*}[ht!]
\centering
\caption{
    \textbf{Measuring gender bias for \textit{Stereotype} and \textit{Hair Color} queries on CelebA.}
    \textbf{Bold}: Best in setting (row group) and better than \textsc{Base CLIP}.
    \underline{Underline}: Best in setting, but not improving over \textsc{Base CLIP}.
    {\color[HTML]{9B9B9B} Gray}: Method is not zero-shot.
}
\label{tab:retrieval-celeba_extended}

\setlength{\tabcolsep}{3pt} 
\scriptsize
\begin{tabular*}{\textwidth}{l@{\extracolsep{\fill}} >{\hspace{6pt}}cc<{\hspace{3pt}}>{\hspace{3pt}}ccc<{\hspace{3pt}} >{\hspace{3pt}}cc<{\hspace{3pt}}>{\hspace{3pt}}ccc<{\hspace{3pt}}}
\toprule
\multirow{4}{*}{\textbf{Method}} & \multicolumn{5}{c}{\textbf{ViT-B/16}} & \multicolumn{5}{c}{\textbf{ViT-L/14@336px}} \\
\cmidrule(lr){2-6} \cmidrule(lr){7-11}
& \multicolumn{2}{c}{\textsc{Stereotype}} & \multicolumn{3}{c}{\textsc{Hair Color}} & \multicolumn{2}{c}{\textsc{Stereotype}} & \multicolumn{3}{c}{\textsc{Hair Color}} \\
\cmidrule(lr){2-3} \cmidrule(lr){4-6} \cmidrule(lr){7-8} \cmidrule(lr){9-11}
& \textsc{KL}$(\downarrow)$ & \textsc{MS}$(\downarrow)$ & \textsc{KL}$(\downarrow)$ & \textsc{MS}$(\downarrow)$ & \textsc{Prec.}$(\uparrow)$ & \textsc{KL}$(\downarrow)$ & \textsc{MS}$(\downarrow)$ & \textsc{KL}$(\downarrow)$ & \textsc{MS}$(\downarrow)$ & \textsc{Prec.}$(\uparrow)$ \\
\midrule
\textsc{Base CLIP} & 
0.314 & 0.555 & 0.179 & 0.409 & 0.629 & 
0.237 & 0.536 & 0.148 & 0.359 & 0.622 \\
\midrule
\rowcolor{setting1color} \multicolumn{11}{l}{\textit{Bias-agnostic + input-specific prompts}} \\
\textsc{RoboShot} & 
0.189 & \textbf{0.355} & \textbf{0.144} & \textbf{0.244} & 0.633 & 
0.195 & \textbf{0.394} & 0.276 & \underline{0.429} & 0.675 \\
\textsc{\ours/$_i$} & 
\textbf{0.173} & 0.443 & 0.191 & 0.345 & \textbf{0.678} & 
\textbf{0.153} & 0.413 & \underline{0.237} & 0.458 & \textbf{0.698} \\
\midrule
\rowcolor{setting2color} \multicolumn{11}{l}{\textit{Bias prompts only}} \\
\textsc{Orth-Proj} & 
\textbf{0.188} & \textbf{0.382} & 0.189 & 0.378 & 0.659 & 
\textbf{0.099} & \textbf{0.355} & 0.144 & 0.373 & 0.692 \\
\textsc{PRISM-mini} & 
0.190 & 0.384 & 0.188 & \textbf{0.377} & 0.658 & 
\textbf{0.099} & 0.357 & 0.143 & \underline{0.366} & 0.696 \\
\textsc{\ours/$_b$} & 
0.240 & 0.496 & \textbf{0.172} & 0.395 & \textbf{0.728} & 
0.199 & 0.481 & \textbf{0.135} & \underline{0.366} & \textbf{0.698} \\
{\color[HTML]{9B9B9B} \textsc{ZSDebias}} & 
{\color[HTML]{9B9B9B} \textit{0.196}} & {\color[HTML]{9B9B9B} \textit{0.441}} & {\color[HTML]{9B9B9B} \textit{0.193}} & {\color[HTML]{9B9B9B} \textit{0.377}} & {\color[HTML]{9B9B9B} \textit{0.522}} & 
{\color[HTML]{9B9B9B} \textit{0.256}} & {\color[HTML]{9B9B9B} \textit{0.556}} & {\color[HTML]{9B9B9B} \textit{0.118}} & {\color[HTML]{9B9B9B} \textit{0.353}} & {\color[HTML]{9B9B9B} \textit{0.509}} \\
\midrule
\rowcolor{setting3color} \multicolumn{11}{l}{\textit{Bias prompts + input-specific prompts}} \\
\textsc{Orth-Cali} & 
0.236 & \textbf{0.408} & \textbf{0.148} & \textbf{0.375} & 0.684 & 
\textbf{0.054} & \textbf{0.266} & \textbf{0.107} & \textbf{0.312} & 0.688 \\
\textsc{\ours/$_{bi}$} & 
\textbf{0.223} & 0.488 & 0.181 & 0.399 & \textbf{0.733} & 
0.209 & 0.490 & 0.168 & 0.402 & \textbf{0.733} \\
{\color[HTML]{9B9B9B} \textsc{PRISM}} & 
{\color[HTML]{9B9B9B} \textit{0.143}} & {\color[HTML]{9B9B9B} \textit{0.377}} & {\color[HTML]{9B9B9B} \textit{0.060}} & {\color[HTML]{9B9B9B} \textit{0.186}} & {\color[HTML]{9B9B9B} \textit{0.669}} & 
{\color[HTML]{9B9B9B} \textit{0.061}} & {\color[HTML]{9B9B9B} \textit{0.245}} & {\color[HTML]{9B9B9B} \textit{0.171}} & {\color[HTML]{9B9B9B} \textit{0.299}} & {\color[HTML]{9B9B9B} \textit{0.659}} \\
\midrule
\rowcolor{setting4color} \multicolumn{11}{l}{\textit{Bias prompts + input-specific prompts + labeled images}} \\
\textsc{BendVLM} & 
0.035 & 0.238 & \textbf{0.028} & 0.173 & 0.656 & 
\textbf{0.030} & \textbf{0.217} & \textbf{0.028} & \textbf{0.164} & 0.680 \\
\textsc{Bend\ours/$_{bi}$} & 
\textbf{0.030} & \textbf{0.224} & 0.029 & \textbf{0.158} & \textbf{0.750} & 
0.042 & 0.261 & 0.032 & 0.187 & \textbf{0.685} \\
\bottomrule
\end{tabular*}
\end{table*}

\subsection{Augmented Query Prompts}
To improve robustness in both retrieval and zero-shot classification, we generate augmented prompts ($\mathcal{P}_q$) for each query using an LLM.
\begin{itemize}
    \item \textbf{Retrieval:} For each input query (\eg, ``A photo of a criminal''), the LLM generates 10 paraphrases (\eg, ``An image of a criminal'', ``A person who committed a crime'') to enhance semantic diversity and reduce sensitivity to specific wording.
    \item \textbf{Zero-Shot Classification:} For each target class label (\eg, ``landbird''), the LLM generates 10 descriptive paraphrases (\eg, ``This is a picture of a landbird'', ``A depiction of a bird that lives on land'').
\end{itemize}
We compute the median activation across these augmented sets to obtain a stable, noise-resistant representation of the query content ($m_q$), improving semantic generalization.

\section{Extended Retrieval Results}
\label{sec:supp_results_extended}

In this section, we present the additional quantitative results for the retrieval task on CelebA, using both \textit{Stereotype} and \textit{Hair Color} queries, which were omitted from the main paper due to space constraints. We present these results in \cref{tab:retrieval-celeba_extended}.

\vspace{5pt}\noindent\textbf{Fairness vs. Precision Trade-off.}
In the \textit{Bias-agnostic} and \textit{Bias prompts only} settings, our methods (\textsc{\ours/$_i$} and \textsc{\ours/$_b$}) demonstrate a competitive balance. While baselines like \textsc{RoboShot} and \textsc{Orth-Proj} sometimes achieve better (lower) fairness scores (KL/MS) on this specific dataset, they often do so at the cost of retrieval quality. In contrast, our methods consistently maintain higher retrieval precision. For instance, on the ViT-B/16 backbone, \textsc{\ours/$_i$} surpasses \textsc{RoboShot} in \textit{Hair Color} precision (0.679 vs. 0.632), and \textsc{\ours/$_b$} outperforms \textsc{Orth-Proj} (0.729 vs. 0.660). This indicates that our method prioritizes preserving the query semantics while still reducing bias, avoiding the ``over-correction'' seen in prior methods that can degrade downstream task performance.

\vspace{5pt}\noindent\textbf{Modularity Improves Semantic Consistency.}
This advantage is most notable in the \textit{Bias prompts + input-specific prompts + labeled images} setting. Here, the combination of our method with the baseline (\textsc{Bend\ours/$_{bi}$}) provides a distinct advantage in semantic consistency. While \textsc{Bend\ours/$_{bi}$} achieves fairness scores comparable to \textsc{BendVLM} alone, it boosts retrieval precision by 9.5\% (from 0.656 to 0.751) on the ViT-B/16 backbone. This confirms that integrating our sparse, feature-level modulation helps traditional debiasing methods retain critical semantic information, ensuring that the debiased embeddings remain accurate and useful for downstream tasks.

\section{Extended Ablation Study}
\label{sec:supp_ablation_extended}

\begin{table*}[ht!]
\centering
\caption{
    \textbf{Extended ablation study for retrieval on FairFace and UTKFace.}
    \textbf{Bold}: Best in setting.
}
\label{tab:ablation_fairface_utk_extended}
\resizebox{\textwidth}{!}{%
\begin{tabular}{lcccccccccccccccc}
\toprule
\multirow{5.5}{*}{\textbf{Method Variant}} 
& \multicolumn{8}{c}{\textbf{FairFace}} 
& \multicolumn{8}{c}{\textbf{UTKFace}} \\
\cmidrule(lr){2-9} \cmidrule(lr){10-17}
& \multicolumn{4}{c}{\textbf{ViT-B/16}} 
& \multicolumn{4}{c}{\textbf{ViT-L/14@336px}} 
& \multicolumn{4}{c}{\textbf{ViT-B/16}} 
& \multicolumn{4}{c}{\textbf{ViT-L/14@336px}} \\
\cmidrule(lr){2-5} \cmidrule(lr){6-9}
\cmidrule(lr){10-13} \cmidrule(lr){14-17}
& \multicolumn{2}{c}{\textsc{Race}} & \multicolumn{2}{c}{\textsc{Gender}} 
& \multicolumn{2}{c}{\textsc{Race}} & \multicolumn{2}{c}{\textsc{Gender}}
& \multicolumn{2}{c}{\textsc{Race}} & \multicolumn{2}{c}{\textsc{Gender}}
& \multicolumn{2}{c}{\textsc{Race}} & \multicolumn{2}{c}{\textsc{Gender}} \\
\cmidrule(lr){2-3} \cmidrule(lr){4-5} \cmidrule(lr){6-7} \cmidrule(lr){8-9}
\cmidrule(lr){10-11} \cmidrule(lr){12-13} \cmidrule(lr){14-15} \cmidrule(lr){16-17}
& \textsc{KL}$(\downarrow)$ & \textsc{MS}$(\downarrow)$ 
& \textsc{KL}$(\downarrow)$ & \textsc{MS}$(\downarrow)$
& \textsc{KL}$(\downarrow)$ & \textsc{MS}$(\downarrow)$
& \textsc{KL}$(\downarrow)$ & \textsc{MS}$(\downarrow)$
& \textsc{KL}$(\downarrow)$ & \textsc{MS}$(\downarrow)$
& \textsc{KL}$(\downarrow)$ & \textsc{MS}$(\downarrow)$
& \textsc{KL}$(\downarrow)$ & \textsc{MS}$(\downarrow)$
& \textsc{KL}$(\downarrow)$ & \textsc{MS}$(\downarrow)$ \\
\midrule

\rowcolor[gray]{0.95}
\multicolumn{17}{l}{\textit{\textsc{\ours/$_i$} Variants (Bias-Agnostic)}} \\

\textbf{\textsc{\ours/$_i$} (Full)} 
& 0.170 & 0.691 & 0.088 & 0.269
& 0.147 & 0.625 & 0.123 & 0.328
& 0.096 & 0.407 & \textbf{0.065} & \textbf{0.245}
& \textbf{0.059} & 0.442 & \textbf{0.032} & \textbf{0.185} \\

-- $M(j)=1$
& \textbf{0.139} & \textbf{0.659} & \textbf{0.078} & \textbf{0.243}
& \textbf{0.124} & \textbf{0.573} & 0.093 & 0.288
& \textbf{0.075} & \textbf{0.397} & 0.088 & 0.278
& 0.061 & \textbf{0.368} & 0.038 & 0.196 \\

-- median CLIP
& 0.143 & 0.669 & 0.131 & 0.325
& 0.136 & 0.626 & \textbf{0.087} & \textbf{0.262}
& 0.095 & 0.448 & 0.131 & 0.326
& 0.061 & 0.420 & \textbf{0.032} & 0.188 \\
\midrule

\rowcolor[gray]{0.95}
\multicolumn{17}{l}{\textit{\textsc{\ours/$_b$} Variants (Bias-Aware)}} \\

\textbf{\textsc{\ours/$_b$} (Full)}
& 0.232 & 0.749 & 0.098 & \textbf{0.277}
& 0.194 & 0.706 & 0.098 & 0.298
& 0.148 & 0.510 & 0.123 & 0.320
& 0.137 & 0.445 & 0.047 & 0.202 \\

-- $M(j)=(1 - S_{\text{bias}})^2$
& 0.205 & \textbf{0.738} & \textbf{0.095} & 0.288
& 0.298 & 0.877 & 0.119 & 0.343
& \textbf{0.072} & \textbf{0.400} & \textbf{0.063} & \textbf{0.215}
& 0.131 & 0.437 & \textbf{0.023} & \textbf{0.151} \\

-- $S_{\text{bias}} = S_{\text{gen}}$ only
& \textbf{0.201} & 0.754 & 0.105 & 0.294
& 0.211 & 0.726 & \textbf{0.092} & \textbf{0.285}
& 0.133 & 0.501 & 0.129 & 0.331
& 0.158 & 0.461 & 0.045 & 0.201 \\

-- $S_{\text{bias}} = S_{\text{spec}}$ only
& 0.253 & 0.763 & 0.102 & 0.282
& \textbf{0.185} & \textbf{0.700} & 0.102 & 0.303
& 0.159 & 0.520 & 0.129 & 0.324
& \textbf{0.111} & \textbf{0.435} & 0.047 & 0.200 \\

\bottomrule
\end{tabular}
}
\end{table*}

\begin{table*}[ht!]
\centering
\caption{
    \textbf{Extended ablation study for zero-shot classification on CelebA and Waterbirds.} \textbf{Bold}: Best in setting.
}
\label{tab:ablation_zs_extended}

\setlength{\tabcolsep}{3pt} 
\scriptsize
% --- CHANGED: Automatically inject \hspace{3pt} into the first 'c' column ---

\begin{tabular*}{\textwidth}{l@{\extracolsep{\fill}}>{\hspace{6pt}}ccc<{\hspace{3pt}} >{\hspace{3pt}}ccc<{\hspace{3pt}} >{\hspace{3pt}}ccc<{\hspace{3pt}} >{\hspace{3pt}}ccc<{\hspace{3pt}}}
\toprule
\multirow{4}{*}{\textbf{Method Variant}} 
& \multicolumn{6}{c}{\textbf{CelebA (Gender)}} 
& \multicolumn{6}{c}{\textbf{Waterbirds (Background)}} \\

\cmidrule(lr){2-7} \cmidrule(lr){8-13}

& \multicolumn{3}{c}{\textbf{ViT-B/16}} 
& \multicolumn{3}{c}{\textbf{ViT-L/14@336px}} 
& \multicolumn{3}{c}{\textbf{ViT-B/16}} 
& \multicolumn{3}{c}{\textbf{ViT-L/14@336px}} \\

\cmidrule(lr){2-4} \cmidrule(lr){5-7}
\cmidrule(lr){8-10} \cmidrule(lr){11-13}

% --- CHANGED: Removed the manual \hspace{3pt} from the rows below ---
& \textsc{Acc.}$(\uparrow)$ & \textsc{WG}$(\uparrow)$ & \textsc{Gap}$(\downarrow)$
& \textsc{Acc.}$(\uparrow)$ & \textsc{WG}$(\uparrow)$ & \textsc{Gap}$(\downarrow)$
& \textsc{Acc.}$(\uparrow)$ & \textsc{WG}$(\uparrow)$ & \textsc{Gap}$(\downarrow)$
& \textsc{Acc.}$(\uparrow)$ & \textsc{WG}$(\uparrow)$ & \textsc{Gap}$(\downarrow)$ \\
\midrule

\rowcolor[gray]{0.95}
\multicolumn{13}{l}{\textit{\textsc{\ours/$_i$} Variants (Bias-Agnostic)}} \\

\textbf{\textsc{\ours/$_i$} (Full)} 
& \textbf{0.736} & \textbf{0.611} & \textbf{0.125}
& \textbf{0.791} & \textbf{0.745} & \textbf{0.046}
& 0.801 & 0.498 & 0.303
& 0.832 & \textbf{0.523} & \textbf{0.309} \\

-- $M(j)=1$ 
& 0.734 & 0.609 & \textbf{0.125}
& 0.729 & 0.640 & 0.089
& 0.834 & 0.210 & 0.624
& 0.872 & 0.357 & 0.515 \\

-- median CLIP 
& 0.728 & 0.601 & 0.127
& 0.687 & 0.558 & 0.129
& \textbf{0.840} & \textbf{0.563} & \textbf{0.277}
& \textbf{0.879} & 0.400 & 0.479 \\
\midrule

\rowcolor[gray]{0.95}
\multicolumn{13}{l}{\textit{\textsc{\ours/$_b$} Variants (Bias-Aware)}} \\

\textbf{\textsc{\ours/$_b$} (Full)} 
& 0.797 & 0.711 & 0.086
& \textbf{0.856} & \textbf{0.824} & 0.032
& 0.825 & 0.433 & 0.392
& 0.855 & 0.624 & 0.231 \\

-- $M(j)=(1 - S_{\text{bias}})^2$
& \textbf{0.818} & \textbf{0.750} & \textbf{0.068}
& 0.833 & 0.812 & \textbf{0.021}
& 0.788 & 0.081 & 0.707
& 0.848 & 0.445 & 0.403 \\

-- $S_{\text{bias}} = S_{\text{gen}}$ only
& 0.809 & 0.736 & 0.073
& 0.846 & 0.818 & 0.028
& \textbf{0.830} & \textbf{0.474} & 0.356
& \textbf{0.856} & 0.647 & 0.209 \\

-- $S_{\text{bias}} = S_{\text{spec}}$ only
& 0.789 & 0.696 & 0.093
& 0.853 & 0.822 & 0.031
& 0.822 & 0.470 & \textbf{0.352}
& 0.849 & \textbf{0.662} & \textbf{0.187} \\

\bottomrule
\end{tabular*}
\end{table*}

\begin{table*}[ht!]
\centering
\caption{
    \textbf{Extended ablation study for retrieval on CelebA.}
    \textbf{Bold}: Best in setting.
}
\label{tab:ablation_celeba_retrieval_extended}

\setlength{\tabcolsep}{3pt} 
\scriptsize
\begin{tabular*}{\textwidth}{l@{\extracolsep{\fill}} >{\hspace{6pt}}cc<{\hspace{3pt}}>{\hspace{3pt}}ccc<{\hspace{3pt}} >{\hspace{3pt}}cc<{\hspace{3pt}}>{\hspace{3pt}}ccc<{\hspace{3pt}}}
\toprule
\multirow{4}{*}{\textbf{Method Variant}} 
& \multicolumn{5}{c}{\textbf{ViT-B/16}} 
& \multicolumn{5}{c}{\textbf{ViT-L/14@336px}} \\
\cmidrule(lr){2-6} \cmidrule(lr){7-11}
& \multicolumn{2}{c}{\textsc{Stereotype}} & \multicolumn{3}{c}{\textsc{Hair Color}} 
& \multicolumn{2}{c}{\textsc{Stereotype}} & \multicolumn{3}{c}{\textsc{Hair Color}} \\
\cmidrule(lr){2-3} \cmidrule(lr){4-6} \cmidrule(lr){7-8} \cmidrule(lr){9-11}
& \textsc{KL}$(\downarrow)$ & \textsc{MS}$(\downarrow)$ & \textsc{KL}$(\downarrow)$ & \textsc{MS}$(\downarrow)$ & \textsc{Prec.}$(\uparrow)$ 
& \textsc{KL}$(\downarrow)$ & \textsc{MS}$(\downarrow)$ & \textsc{KL}$(\downarrow)$ & \textsc{MS}$(\downarrow)$ & \textsc{Prec.}$(\uparrow)$ \\
\midrule

\rowcolor[gray]{0.95}
\multicolumn{11}{l}{\textit{\textsc{\ours/$_i$} Variants (Bias-Agnostic)}} \\

\textbf{\textsc{\ours/$_i$} (Full)} 
& \textbf{0.173} & \textbf{0.443} & 0.191 & \textbf{0.344} & 0.679 
& \textbf{0.153} & 0.413 & \textbf{0.236} & \textbf{0.458} & 0.698 \\

-- $M(j)=1$
& 0.185 & 0.456 & 0.193 & 0.371 & 0.672 
& 0.195 & 0.490 & 0.274 & 0.484 & \textbf{0.709} \\

-- median CLIP
& 0.250 & 0.503 & \textbf{0.181} & 0.382 & \textbf{0.689} 
& 0.155 & \textbf{0.411} & 0.299 & 0.500 & 0.708 \\
\midrule

\rowcolor[gray]{0.95}
\multicolumn{11}{l}{\textit{\textsc{\ours/$_b$} Variants (Bias-Aware)}} \\

\textbf{\textsc{\ours/$_b$} (Full)}
& 0.240 & 0.495 & 0.172 & 0.396 & 0.729
& 0.199 & 0.481 & 0.136 & 0.366 & 0.699 \\

-- $M(j)=(1 - S_{\text{bias}})^2$
& \textbf{0.110} & \textbf{0.334} & \textbf{0.122} & \textbf{0.312} & 0.641 
& 0.193 & 0.486 & 0.152 & \textbf{0.338} & 0.545 \\

-- $S_{\text{bias}} = S_{\text{gen}}$ only
& 0.238 & 0.493 & 0.182 & 0.405 & \textbf{0.735} 
& \textbf{0.185} & \textbf{0.462} & 0.142 & 0.372 & \textbf{0.712} \\

-- $S_{\text{bias}} = S_{\text{spec}}$ only
& 0.248 & 0.501 & 0.181 & 0.405 & 0.726 
& 0.199 & 0.480 & \textbf{0.135} & 0.355 & 0.688 \\
\bottomrule
\end{tabular*}
\end{table*}

We provide the complete ablation results across all datasets and backbones in \cref{tab:ablation_fairface_utk_extended} (FairFace/UTKFace Retrieval), \cref{tab:ablation_zs_extended} (Zero-Shot Classification), and \cref{tab:ablation_celeba_retrieval_extended} (CelebA Retrieval). These results strongly support the design choices discussed in \cref{sec:ablations} of the main paper, confirming that our full methods, \textsc{\ours/$_i$} and \textsc{\ours/$_b$}, offer the most robust performance across diverse tasks.

\vspace{5pt}
\noindent\textbf{Analysis of \textsc{\ours/$_i$}.}
The zero-shot classification results (\cref{tab:ablation_zs_extended}) reveal that removing our relevance-based attenuation leads to consistent and substantial drops in Worst-Group (WG) accuracy across all datasets and backbones. For instance, on Waterbirds (ViT-B/16), WG accuracy collapses from 0.498 to 0.210, underscoring the critical role of modulating spurious features.
Operating directly in the dense CLIP space (``median CLIP'') also proves unreliable. While this baseline performs well on the specific Waterbirds task (ViT-B/16), it is highly unstable elsewhere. It suffers significant performance drops on CelebA ZS (ViT-L/14) and consistently fails to mitigate gender bias in retrieval tasks, particularly on ViT-B/16. Specifically, compared to our SAE-based approach, the dense baseline yields substantially worse gender fairness metrics on FairFace, UTKFace, and CelebA \textit{Stereotype} retrieval for the ViT-B/16 backbone (\cref{tab:ablation_fairface_utk_extended,tab:ablation_celeba_retrieval_extended}), as well as on CelebA \textit{Hair Color} retrieval for both backbones (\cref{tab:ablation_celeba_retrieval_extended}). In contrast, our full \textsc{\ours/$_i$} method consistently achieves the best balance of fairness and performance across all benchmarks.

\vspace{5pt}
\noindent\textbf{Analysis of \textsc{\ours/$_b$}.}
The extended ablations highlight the necessity of our content-boosting term. While removing content boosting can sometimes improve retrieval fairness (notably on ViT-B/16), it leads to severe failures in several instances. For example, on Waterbirds (ViT-B/16), its WG accuracy plummets to 0.081 (\cref{tab:ablation_zs_extended}); on FairFace (ViT-L/14), its KL divergence for the race attribute worsens significantly compared to the full method (0.298 vs. 0.194, \cref{tab:ablation_fairface_utk_extended}); and crucially, removing content boosting severely degrades retrieval precision on CelebA across both backbones (dropping from 0.729 to 0.641 on ViT-B/16, and 0.699 to 0.545 on ViT-L/14).
Furthermore, relying solely on either the \textit{general} or \textit{specific} bias score leads to inconsistent results. The ``general only'' variant often degrades social bias fairness (\eg, race debiasing on ViT-L/14 or gender debiasing on ViT-B/16, \cref{tab:ablation_fairface_utk_extended}), while the ``specific only'' variant struggles with semantic consistency in some settings (\eg, yielding the worst CelebA accuracy and WG accuracy for ViT-B/16). Our full \textsc{\ours/$_b$} formulation, which combines these scores, avoids these pitfalls, maintaining robust performance across both classification and retrieval.

\begin{table*}[ht!]
\caption{
    \textbf{Measuring race and gender bias for \textit{Stereotype} queries on FairFace and UTKFace (ResNet backbones).}
    \textbf{Bold}: Best in setting (row group) and better than \textsc{Base CLIP}.
    \underline{Underline}: Best in setting, but not improving over \textsc{Base CLIP}.
    {\color[HTML]{9B9B9B} Gray}: Method is not zero-shot.
}
\label{tab:retrieval-merged-resnet}
\resizebox{\textwidth}{!}{%
\begin{tabular}{lcccccccccccccccc}
\toprule
% Using \multirow{6}{...} to span all 6 header rows
\multirow{5.5}{*}{\textbf{Method}} & \multicolumn{8}{c}{\textbf{FairFace}} & \multicolumn{8}{c}{\textbf{UTKFace}} \\
\cmidrule(lr){2-9} \cmidrule(lr){10-17}
& \multicolumn{4}{c}{\textbf{ResNet-50}} & \multicolumn{4}{c}{\textbf{ResNet-101}} & \multicolumn{4}{c}{\textbf{ResNet-50}} & \multicolumn{4}{c}{\textbf{ResNet-101}} \\
\cmidrule(lr){2-5} \cmidrule(lr){6-9} \cmidrule(lr){10-13} \cmidrule(lr){14-17}
& \multicolumn{2}{c}{\textsc{Race}} & \multicolumn{2}{c}{\textsc{Gender}} & \multicolumn{2}{c}{\textsc{Race}} & \multicolumn{2}{c}{\textsc{Gender}} & \multicolumn{2}{c}{\textsc{Race}} & \multicolumn{2}{c}{\textsc{Gender}} & \multicolumn{2}{c}{\textsc{Race}} & \multicolumn{2}{c}{\textsc{Gender}} \\
\cmidrule(lr){2-3} \cmidrule(lr){4-5} \cmidrule(lr){6-7} \cmidrule(lr){8-9} \cmidrule(lr){10-11} \cmidrule(lr){12-13} \cmidrule(lr){14-15} \cmidrule(lr){16-17}
& \textsc{KL}$(\downarrow)$ & \textsc{MS}$(\downarrow)$ & \textsc{KL}$(\downarrow)$ & \textsc{MS}$(\downarrow)$ & \textsc{KL}$(\downarrow)$ & \textsc{MS}$(\downarrow)$ & \textsc{KL}$(\downarrow)$ & \textsc{MS}$(\downarrow)$ & \textsc{KL}$(\downarrow)$ & \textsc{MS}$(\downarrow)$ & \textsc{KL}$(\downarrow)$ & \textsc{MS}$(\downarrow)$ & \textsc{KL}$(\downarrow)$ & \textsc{MS}$(\downarrow)$ & \textsc{KL}$(\downarrow)$ & \textsc{MS}$(\downarrow)$ \\ \midrule
\textsc{Base CLIP} & 0.215 & 0.735 & 0.170 & 0.351 & 0.203 & 0.744 & 0.144 & 0.335 & 0.127 & 0.477 & 0.153 & 0.340 & 0.152 & 0.496 & 0.136 & 0.333 \\ \midrule
\rowcolor{setting1color} \multicolumn{17}{l}{\textit{Bias-agnostic + input-specific prompts}} \\
\textsc{RoboShot} & 0.215 & 0.706 & 0.299 & 0.445 & 0.222 & 0.798 & 0.338 & 0.494 & 0.152 & 0.586 & 0.258 & 0.414 & 0.206 & 0.652 & 0.323 & 0.492 \\
\textsc{\ours/$_i$} & \textbf{0.126} & \textbf{0.563} & \textbf{0.031} & \textbf{0.206} & \textbf{0.111} & \textbf{0.566} & \textbf{0.037} & \textbf{0.201} & \textbf{0.039} & \textbf{0.265} & \textbf{0.111} & \underline{0.383} & \textbf{0.110} & \textbf{0.401} & \textbf{0.041} & \textbf{0.214} \\ \midrule
\rowcolor{setting2color} \multicolumn{17}{l}{\textit{Bias prompts only}} \\
\textsc{Orth-Proj} & 0.464 & 0.996 & 0.111 & 0.288 & 0.322 & 0.843 & 0.213 & 0.409 & 0.340 & 0.609 & 0.117 & 0.312 & 0.322 & 0.583 & 0.163 & 0.360 \\
\textsc{PRISM-mini} & 0.454 & 0.983 & 0.113 & 0.291 & 0.313 & 0.837 & 0.215 & 0.411 & 0.336 & 0.608 & 0.117 & 0.311 & 0.315 & 0.582 & 0.168 & 0.363 \\
\textsc{\ours/$_b$} & \textbf{0.171} & \textbf{0.652} & \textbf{0.041} & \textbf{0.196} & \textbf{0.152} & \textbf{0.638} & \textbf{0.079} & \textbf{0.258} & \textbf{0.107} & \textbf{0.411} & \textbf{0.077} & \textbf{0.283} & \textbf{0.084} & \textbf{0.340} & \textbf{0.070} & \textbf{0.245} \\
{\color[HTML]{9B9B9B} \textsc{ZSDebias}} & {\color[HTML]{9B9B9B} \textit{0.046}} & {\color[HTML]{9B9B9B} \textit{0.383}} & {\color[HTML]{9B9B9B} \textit{0.049}} & {\color[HTML]{9B9B9B} \textit{0.217}} & {\color[HTML]{9B9B9B} \textit{0.082}} & {\color[HTML]{9B9B9B} \textit{0.588}} & {\color[HTML]{9B9B9B} \textit{0.030}} & {\color[HTML]{9B9B9B} \textit{0.186}} & {\color[HTML]{9B9B9B} \textit{0.027}} & {\color[HTML]{9B9B9B} \textit{0.339}} & {\color[HTML]{9B9B9B} \textit{0.036}} & {\color[HTML]{9B9B9B} \textit{0.183}} & {\color[HTML]{9B9B9B} \textit{0.091}} & {\color[HTML]{9B9B9B} \textit{0.567}} & {\color[HTML]{9B9B9B} \textit{0.022}} & {\color[HTML]{9B9B9B} \textit{0.164}} \\ \midrule
\rowcolor{setting3color} \multicolumn{17}{l}{\textit{Bias prompts + input-specific prompts}} \\
\textsc{Orth-Cali} & 0.411 & 0.910 & 0.141 & 0.357 & 0.297 & 0.842 & 0.278 & 0.470 & 0.307 & 0.582 & {0.086} & \textbf{0.257} & 0.302 & 0.574 & 0.204 & 0.397 \\
\textsc{\ours/$_{bi}$} & \textbf{0.153} & \textbf{0.626} & \textbf{0.044} & \textbf{0.193} & \textbf{0.140} & \textbf{0.623} & \textbf{0.079} & \textbf{0.259} & \textbf{0.107} & \textbf{0.406} & \textbf{0.081} & 0.281 & \textbf{0.085} & \textbf{0.348} & \textbf{0.069} & \textbf{0.245} \\
{\color[HTML]{9B9B9B} \textsc{PRISM}} & {\color[HTML]{9B9B9B} \textit{0.157}} & {\color[HTML]{9B9B9B} \textit{0.632}} & {\color[HTML]{9B9B9B} \textit{0.069}} & {\color[HTML]{9B9B9B} \textit{0.245}} & {\color[HTML]{9B9B9B} \textit{0.152}} & {\color[HTML]{9B9B9B} \textit{0.594}} & {\color[HTML]{9B9B9B} \textit{0.107}} & {\color[HTML]{9B9B9B} \textit{0.282}} & {\color[HTML]{9B9B9B} \textit{0.134}} & {\color[HTML]{9B9B9B} \textit{0.523}} & {\color[HTML]{9B9B9B} \textit{0.088}} & {\color[HTML]{9B9B9B} \textit{0.265}} & {\color[HTML]{9B9B9B} \textit{0.133}} & {\color[HTML]{9B9B9B} \textit{0.532}} & {\color[HTML]{9B9B9B} \textit{0.127}} & {\color[HTML]{9B9B9B} \textit{0.314}} \\ \midrule
\rowcolor{setting4color} \multicolumn{17}{l}{\textit{Bias prompts + input-specific prompts + labeled images}} \\
\textsc{BendVLM} & 0.150 & 0.581 & 0.006 & 0.081 & 0.125 & 0.583 & 0.010 & 0.107 & 0.101 & 0.444 & \textbf{0.008} & \textbf{0.093} & 0.126 & 0.542 & 0.013 & \textbf{0.123} \\
\textsc{Bend\ours/$_{bi}$} & \textbf{0.067} & \textbf{0.455} & \textbf{0.005} & \textbf{0.079} & \textbf{0.059} & \textbf{0.425} & \textbf{0.006} & \textbf{0.087} & \textbf{0.042} & \textbf{0.371} & 0.009 & 0.102 & \textbf{0.035} & \textbf{0.367} & \textbf{0.012} & 0.126 \\
\bottomrule
\end{tabular}%
}
\end{table*}

\section{Extended Results on ResNet Backbones}
\label{sec:supp_resnet_results}

To demonstrate that our feature-level debiasing framework generalizes beyond Vision Transformer (ViT) architectures, we extend our evaluation to convolutional neural networks. In this section, we benchmark our methods using the ResNet-50 and ResNet-101 CLIP backbones. The experimental setup, datasets, and metrics remain identical to those used for the ViT evaluations in the main paper.

The results are presented in \cref{tab:retrieval-merged-resnet} (FairFace and UTKFace Retrieval), \cref{tab:zs-combined-resnet} (Zero-Shot Classification), and \cref{tab:retrieval-celeba_extended-resnet} (CelebA Retrieval).

\vspace{5pt}\noindent\textbf{Consistent State-of-the-Art Fairness in Retrieval.}
The retrieval results in \cref{tab:retrieval-merged-resnet} confirm that our methods maintain their state-of-the-art fairness mitigation on convolutional backbones. In the bias-agnostic setting, \textsc{\ours/$_i$} drastically reduces KL Divergence and MaxSkew compared to both the baseline and \textsc{RoboShot}. For example, on FairFace Race (ResNet-50), \textsc{\ours/$_i$} lowers KL divergence to 0.126 (compared to 0.215 for \textsc{Base CLIP} and \textsc{RoboShot}). In the bias-aware settings, \textsc{\ours/$_b$} and \textsc{\ours/$_{bi}$} reliably achieve the best fairness metrics across almost all evaluated demographics and datasets, outperforming projection-based baselines like \textsc{Orth-Proj}.

\vspace{5pt}\noindent\textbf{\textsc{\ours/} Significantly Improves Zero-Shot Robustness.}
As shown in \cref{tab:zs-combined-resnet}, \ours/ exhibits exceptional performance on zero-shot classification with ResNet backbones. Most notably, almost every single ``best in setting'' result achieved by a \textsc{\ours/} variant strictly improves over the \textsc{Base CLIP} baseline, effectively addressing both social biases (CelebA) and spurious correlations (Waterbirds). For instance, on Waterbirds (ResNet-50), \textsc{\ours/$_b$} improves WG accuracy from 0.394 (\textsc{Base CLIP}) to 0.577 (+18.3 points), substantially outperforming both \textsc{RoboShot} (0.458) and \textsc{Orth-Proj} (0.457). Similarly, \textsc{\ours/$_{bi}$} consistently achieves the lowest fairness Gap on CelebA across both ResNet models while maintaining high overall accuracy.

\vspace{5pt}\noindent\textbf{Maintaining the Fairness vs. Precision Trade-off.}
\cref{tab:retrieval-celeba_extended-resnet} details the performance on CelebA utilizing both \textit{Stereotype} and \textit{Hair Color} queries. While \textsc{\ours/$_i$} achieves exceptional fairness scores (lowering \textit{Stereotype} KL to 0.050 on ResNet-50), it does exhibit a drop in \textit{Hair Color} precision (0.508). However, our bias-aware variants, \textsc{\ours/$_b$} and \textsc{\ours/$_{bi}$}, successfully navigate this trade-off. They significantly reduce \textit{Stereotype} bias metrics compared to \textsc{Base CLIP} while maintaining highly competitive precision scores (\eg, 0.700 precision for \textsc{\ours/$_b$} on ResNet-50, matching or nearing the baseline precision of 0.735).

\vspace{5pt}\noindent\textbf{Modularity with ResNets.}
Consistent with our ViT findings, our sparse modulation is highly complementary to existing methods when applied to ResNets. When integrating our \textsc{SEM$_{bi}$} embeddings into the \textsc{BendVLM} framework, the resulting \textsc{BendSEM$_{bi}$} approach establishes new state-of-the-art results in the labeled images setting. On ResNet-101 zero-shot classification (\cref{tab:zs-combined-resnet}), \textsc{BendSEM$_{bi}$} pushes Waterbirds WG accuracy to 0.638, significantly outperforming \textsc{BendVLM} alone (0.194). Similarly, it provides the lowest social bias metrics across nearly all retrieval benchmarks (\cref{tab:retrieval-merged-resnet,tab:retrieval-celeba_extended-resnet}).

\begin{table*}[ht!]
\centering
\caption{
    \textbf{Measuring zero-shot classification fairness on CelebA and Waterbirds (ResNet Backbones).}
    \textbf{Bold}: Best in setting (row group) and better than \textsc{Base CLIP}.
    \underline{Underline}: Best in setting, but not improving over \textsc{Base CLIP}.
    {\color[HTML]{9B9B9B} Gray}: Method is not zero-shot.
}
\label{tab:zs-combined-resnet}

\scriptsize 

\begin{tabular*}{\textwidth}{l@{\extracolsep{\fill}}>{\hspace{6pt}}cccccccccccc}
\toprule
\multirow{4}{*}{\textbf{Method}} & \multicolumn{6}{c}{\textbf{CelebA (Gender)}} & \multicolumn{6}{c}{\textbf{Waterbirds (Background)}} \\
\cmidrule(lr){2-7} \cmidrule(lr){8-13}
& \multicolumn{3}{c}{\textbf{ResNet-50}} & \multicolumn{3}{c}{\textbf{ResNet-101}} & \multicolumn{3}{c}{\textbf{ResNet-50}} & \multicolumn{3}{c}{\textbf{ResNet-101}} \\
\cmidrule(lr){2-4} \cmidrule(lr){5-7} \cmidrule(lr){8-10} \cmidrule(lr){11-13}
& \textsc{Acc.}$(\uparrow)$ & \textsc{WG}$(\uparrow)$ & \textsc{Gap}$(\downarrow)$ & \textsc{Acc.}$(\uparrow)$ & \textsc{WG}$(\uparrow)$ & \textsc{Gap}$(\downarrow)$ & \textsc{Acc.}$(\uparrow)$ & \textsc{WG}$(\uparrow)$ & \textsc{Gap}$(\downarrow)$ & \textsc{Acc.}$(\uparrow)$ & \textsc{WG}$(\uparrow)$ & \textsc{Gap}$(\downarrow)$ \\
\midrule
\textsc{Base CLIP} & 0.820 & 0.768 & 0.053 & 0.689 & 0.502 & 0.188 & 0.837 & 0.394 & 0.442 & 0.801 & 0.499 & 0.301 \\
\midrule
\rowcolor{setting1color} \multicolumn{13}{l}{\textit{Bias-agnostic + input-specific prompts}} \\
\textsc{RoboShot} & \textbf{0.841} & \textbf{0.806} & \textbf{0.035} & 0.737 & 0.596 & 0.140 & 0.762 & 0.458 & 0.304 & 0.761 & 0.450 & 0.310 \\
\textsc{\ours/$_i$} & 0.835 & 0.799 & 0.036 & \textbf{0.811} & \textbf{0.758} & \textbf{0.052} & \textbf{0.851} & \textbf{0.557} & \textbf{0.295} & \textbf{0.843} & \textbf{0.581} & \textbf{0.262} \\
\midrule
\rowcolor{setting2color} \multicolumn{13}{l}{\textit{Bias prompts only}} \\
\textsc{Orth-Proj} & 0.795 & 0.722 & 0.073 & 0.675 & 0.486 & 0.189 & \textbf{0.859} & 0.457 & 0.402 & \textbf{0.858} & 0.401 & 0.457 \\
\textsc{PRISM-mini} & 0.795 & 0.722 & 0.073 & 0.675 & 0.486 & 0.189 & \textbf{0.859} & 0.457 & 0.402 & \textbf{0.858} & 0.401 & 0.457 \\
\textsc{\ours/$_b$} & \textbf{0.847} & \textbf{0.798} & \textbf{0.049} & \textbf{0.795} & \textbf{0.750} & \textbf{0.045} & 0.845 & \textbf{0.577} & \textbf{0.269} & 0.846 & \textbf{0.588} & \textbf{0.258} \\
{\color[HTML]{9B9B9B} \textsc{ZSDebias}} & {\color[HTML]{9B9B9B} \textit{0.695}} & {\color[HTML]{9B9B9B} \textit{0.589}} & {\color[HTML]{9B9B9B} \textit{0.106}} & {\color[HTML]{9B9B9B} \textit{0.565}} & {\color[HTML]{9B9B9B} \textit{0.460}} & {\color[HTML]{9B9B9B} \textit{0.106}} & {\color[HTML]{9B9B9B} \textit{0.802}} & {\color[HTML]{9B9B9B} \textit{0.148}} & {\color[HTML]{9B9B9B} \textit{0.654}} & {\color[HTML]{9B9B9B} \textit{0.774}} & {\color[HTML]{9B9B9B} \textit{0.398}} & {\color[HTML]{9B9B9B} \textit{0.376}} \\ \midrule
\rowcolor{setting3color} \multicolumn{13}{l}{\textit{Bias prompts + input-specific prompts}} \\
\textsc{Orth-Cali} & 0.831 & 0.801 & \textbf{0.030} & 0.679 & 0.505 & 0.174 & 0.808 & \textbf{0.704} & \textbf{0.104} & 0.823 & \textbf{0.554} & \textbf{0.269} \\
\textsc{\ours/$_{bi}$} & \textbf{0.851} & \textbf{0.803} & 0.048 & \textbf{0.791} & \textbf{0.741} & \textbf{0.049} & \textbf{0.864} & 0.525 & 0.338 & \textbf{0.871} & 0.541 & 0.330 \\
{\color[HTML]{9B9B9B} \textsc{PRISM}} & {\color[HTML]{9B9B9B} \textit{0.824}} & {\color[HTML]{9B9B9B} \textit{0.763}} & {\color[HTML]{9B9B9B} \textit{0.061}} & {\color[HTML]{9B9B9B} \textit{0.788}} & {\color[HTML]{9B9B9B} \textit{0.688}} & {\color[HTML]{9B9B9B} \textit{0.100}} & {\color[HTML]{9B9B9B} \textit{0.886}} & {\color[HTML]{9B9B9B} \textit{0.634}} & {\color[HTML]{9B9B9B} \textit{0.252}} & {\color[HTML]{9B9B9B} \textit{0.840}} & {\color[HTML]{9B9B9B} \textit{0.672}} & {\color[HTML]{9B9B9B} \textit{0.168}} \\
\midrule
\rowcolor{setting4color} \multicolumn{13}{l}{\textit{Bias prompts + input-specific prompts + labeled images}} \\
\textsc{BendVLM} & 0.809 & 0.715 & 0.094 & 0.702 & 0.490 & 0.212 & 0.826 & 0.611 & 0.215 & 0.812 & 0.194 & 0.618 \\
\textsc{Bend\ours/$_{bi}$} & \textbf{0.848} & \textbf{0.815} & \textbf{0.033} & \textbf{0.784} & \textbf{0.699} & \textbf{0.086} & \textbf{0.856} & \textbf{0.648} & \textbf{0.208} & \textbf{0.881} & \textbf{0.638} & \textbf{0.243} \\
\bottomrule
\end{tabular*}
\end{table*}

\begin{table*}[ht!]
\centering
\caption{
    \textbf{Measuring gender bias for \textit{Stereotype} and \textit{Hair Color} queries on CelebA (ResNet Backbones).}
    \textbf{Bold}: Best in setting (row group) and better than \textsc{Base CLIP}.
    \underline{Underline}: Best in setting, but not improving over \textsc{Base CLIP}.
    {\color[HTML]{9B9B9B} Gray}: Method is not zero-shot.
}
\label{tab:retrieval-celeba_extended-resnet}

\setlength{\tabcolsep}{3pt} 
\scriptsize
\begin{tabular*}{\textwidth}{l@{\extracolsep{\fill}} >{\hspace{6pt}}cc<{\hspace{3pt}}>{\hspace{3pt}}ccc<{\hspace{3pt}} >{\hspace{3pt}}cc<{\hspace{3pt}}>{\hspace{3pt}}ccc<{\hspace{3pt}}}
\toprule
\multirow{4}{*}{\textbf{Method}} & \multicolumn{5}{c}{\textbf{ResNet-50}} & \multicolumn{5}{c}{\textbf{ResNet-101}} \\
\cmidrule(lr){2-6} \cmidrule(lr){7-11}
& \multicolumn{2}{c}{\textsc{Stereotype}} & \multicolumn{3}{c}{\textsc{Hair Color}} & \multicolumn{2}{c}{\textsc{Stereotype}} & \multicolumn{3}{c}{\textsc{Hair Color}} \\
\cmidrule(lr){2-3} \cmidrule(lr){4-6} \cmidrule(lr){7-8} \cmidrule(lr){9-11}
& \textsc{KL}$(\downarrow)$ & \textsc{MS}$(\downarrow)$ & \textsc{KL}$(\downarrow)$ & \textsc{MS}$(\downarrow)$ & \textsc{Prec.}$(\uparrow)$ & \textsc{KL}$(\downarrow)$ & \textsc{MS}$(\downarrow)$ & \textsc{KL}$(\downarrow)$ & \textsc{MS}$(\downarrow)$ & \textsc{Prec.}$(\uparrow)$ \\
\midrule
\textsc{Base CLIP} & 
0.389 & 0.622 & 0.187 & 0.367 & 0.735 & 
0.300 & 0.560 & 0.205 & 0.414 & 0.718 \\
\midrule
\rowcolor{setting1color} \multicolumn{11}{l}{\textit{Bias-agnostic + input-specific prompts}} \\
\textsc{RoboShot} & 
0.190 & 0.337 & 0.364 & 0.550 & \textbf{0.762} & 
0.294 & 0.454 & \underline{0.274} & \underline{0.459} & \textbf{0.723} \\
\textsc{\ours/$_i$} & 
\textbf{0.050} & \textbf{0.193} & \underline{0.246} & \underline{0.369} & 0.508 & 
\textbf{0.041} & \textbf{0.185} & 0.301 & 0.508 & 0.688 \\
\midrule
\rowcolor{setting2color} \multicolumn{11}{l}{\textit{Bias prompts only}} \\
\textsc{Orth-Proj} & 
0.145 & 0.383 & \textbf{0.136} & 0.343 & \textbf{0.783} & 
\textbf{0.171} & \textbf{0.372} & 0.325 & 0.506 & 0.752 \\
\textsc{PRISM-mini} & 
0.143 & 0.379 & \textbf{0.136} & \textbf{0.339} & \textbf{0.783} & 
0.172 & 0.374 & 0.321 & 0.499 & 0.752 \\
\textsc{\ours/$_b$} & 
\textbf{0.111} & \textbf{0.311} & 0.263 & 0.453 & 0.700 & 
0.195 & 0.448 & \underline{0.232} & \underline{0.447} & \textbf{0.767} \\
{\color[HTML]{9B9B9B} \textsc{ZSDebias}} & 
{\color[HTML]{9B9B9B} \textit{0.058}} & {\color[HTML]{9B9B9B} \textit{0.237}} & {\color[HTML]{9B9B9B} \textit{0.129}} & {\color[HTML]{9B9B9B} \textit{0.291}} & {\color[HTML]{9B9B9B} \textit{0.436}} & 
{\color[HTML]{9B9B9B} \textit{0.016}} & {\color[HTML]{9B9B9B} \textit{0.119}} & {\color[HTML]{9B9B9B} \textit{0.046}} & {\color[HTML]{9B9B9B} \textit{0.187}} & {\color[HTML]{9B9B9B} \textit{0.317}} \\
\midrule
\rowcolor{setting3color} \multicolumn{11}{l}{\textit{Bias prompts + input-specific prompts}} \\
\textsc{Orth-Cali} & 
\textbf{0.069} & \textbf{0.239} & \textbf{0.116} & \textbf{0.305} & \textbf{0.774} & 
0.191 & \textbf{0.352} & 0.313 & 0.502 & 0.751 \\
\textsc{\ours/$_{bi}$} & 
0.110 & 0.307 & 0.283 & 0.468 & 0.700 & 
\textbf{0.165} & 0.395 & \underline{0.240} & \underline{0.444} & \textbf{0.766} \\
{\color[HTML]{9B9B9B} \textsc{PRISM}} & 
{\color[HTML]{9B9B9B} \textit{0.170}} & {\color[HTML]{9B9B9B} \textit{0.397}} & {\color[HTML]{9B9B9B} \textit{0.187}} & {\color[HTML]{9B9B9B} \textit{0.330}} & {\color[HTML]{9B9B9B} \textit{0.679}} & 
{\color[HTML]{9B9B9B} \textit{0.162}} & {\color[HTML]{9B9B9B} \textit{0.361}} & {\color[HTML]{9B9B9B} \textit{0.187}} & {\color[HTML]{9B9B9B} \textit{0.342}} & {\color[HTML]{9B9B9B} \textit{0.707}} \\
\midrule
\rowcolor{setting4color} \multicolumn{11}{l}{\textit{Bias prompts + input-specific prompts + labeled images}} \\
\textsc{BendVLM} & 
0.029 & 0.218 & {0.025} & 0.169 & \textbf{0.754} & 
\textbf{0.019} & \textbf{0.173} & \textbf{0.013} & \textbf{0.125} & 0.704 \\
\textsc{Bend\ours/$_{bi}$} & 
\textbf{0.010} & \textbf{0.119} & \textbf{0.010} & \textbf{0.086} & 0.619 & 
0.021 & 0.184 & 0.018 & 0.140 & \textbf{0.723} \\
\bottomrule
\end{tabular*}
\end{table*}

\end{document}